\pdfoutput=1
\documentclass{article}

\usepackage{microtype}
\usepackage{graphicx}
\usepackage{subcaption}
\usepackage{wrapfig}
\usepackage{booktabs}
\PassOptionsToPackage{hyphens}{url}\usepackage{hyperref}
\usepackage{url}
\usepackage{amsfonts}
\usepackage{nicefrac}
\usepackage{pifont}

\usepackage{lipsum}
\usepackage{tabularx}
\usepackage{rotating}

\usepackage[inline]{enumitem}

\usepackage[most]{tcolorbox}
\usepackage{graphicx}
\usepackage[T1]{fontenc}
\usepackage{xcolor}
\hypersetup{
    colorlinks,
    linkcolor={red!50!black},
    citecolor={blue!50!black},
    urlcolor={blue!80!black}
}
\usepackage{cleveref}
\usepackage{alltt}
\usepackage{suffix}
\usepackage{longtable}
\usepackage{array}
\usepackage{capt-of}

\PassOptionsToPackage{numbers, compress}{natbib}
\usepackage[final]{include/neurips_2020}
\newcounter{trCounter}
\newif\iftrvar
\trvartrue
\iftrvar
\newcommand{\tim}[1]{{\small \color{red} \refstepcounter{trCounter}\textsf{[TR]$_{\arabic{trCounter}}$:{#1}}}}

\newcommand{\heiner}[1]{{\small \color{green!50!black} \refstepcounter{trCounter}\textsf{[HK]$_{\arabic{trCounter}}$:{#1}}}}
\newcommand{\nn}[1]{{\small \color{purple} \refstepcounter{trCounter}\textsf{[NN]$_{\arabic{trCounter}}$:{#1}}}}
\newcommand{\roberta}[1]{{\small \color{blue} \refstepcounter{trCounter}\textsf{[RR]$_{\arabic{trCounter}}$:{#1}}}}
\newcommand{\ed}[1]{{\small \color{magenta} \refstepcounter{trCounter}\textsf{[ETG]$_{\arabic{trCounter}}$:{#1}}}}
\newcommand{\alex}[1]{{\small \color{olive} \refstepcounter{trCounter}\textsf{[AHM]$_{\arabic{trCounter}}$:{#1}}}}

\else
\newcommand{\tim}[1]{}

\newcommand{\heiner}[1]{}
\newcommand{\nn}[1]{}
\newcommand{\roberta}[1]{}
\newcommand{\ed}[1]{}
\newcommand{\alex}[1]{}

\fi

\newcommand{\nethack}{NetHack}
\newcommand{\hack}{Hack}
\newcommand{\rogue}{Rogue}
\newcommand{\dcss}{Dungeon Crawl Stone Soup}
\newcommand{\nethackenv}{\texttt{NetHack Learning Environment}}
\newcommand{\NLE}{\texttt{NLE}}

\defcitealias{berlin}{\scshape IRDC, 2008}

\usepackage{soul}
\definecolor{nethack_black}{RGB}{0, 0, 0}
\definecolor{nethack_red}{RGB}{172, 33, 0}
\definecolor{nethack_green}{RGB}{109, 193, 45}
\definecolor{nethack_yellow}{RGB}{196, 194, 45}
\definecolor{nethack_blue}{RGB}{24, 41, 194}
\definecolor{nethack_magenta}{RGB}{173, 51, 194}
\definecolor{nethack_cyan}{RGB}{110, 195, 197}
\definecolor{nethack_light_gray}{RGB}{198, 198, 198}
\definecolor{nethack_gray}{RGB}{104, 104, 104}
\definecolor{nethack_orange}{RGB}{225, 110, 103}
\definecolor{nethack_light_green}{RGB}{159, 250, 115}
\definecolor{nethack_light_yellow}{RGB}{254, 252, 114}
\definecolor{nethack_light_blue}{RGB}{107, 113, 251}
\definecolor{nethack_light_magenta}{RGB}{226, 118, 251}
\definecolor{nethack_light_cyan}{RGB}{161, 252, 255}
\definecolor{nethack_white}{RGB}{255, 255, 255}

\newcommand{\AcidBlob}{{\bf\ttfamily\color{nethack_green}\sethlcolor{black}\hl{b}}}

\newcommand{\Coyote}{{\bf\ttfamily\color{nethack_yellow}\sethlcolor{black}\hl{d}}}

\newcommand{\LittleDog}{{\bf\ttfamily\color{nethack_white}\sethlcolor{black}\hl{d}}}

\newcommand{\Wolf}{{\bf\ttfamily\color{nethack_yellow}\sethlcolor{black}\hl{d}}}

\newcommand{\HellHound}{{\bf\ttfamily\color{nethack_red}\sethlcolor{black}\hl{d}}}

\newcommand{\Kitten}{{\bf\ttfamily\color{nethack_white}\sethlcolor{black}\hl{f}}}

\newcommand{\MindFlayer}{{\bf\ttfamily\color{nethack_magenta}\sethlcolor{black}\hl{h}}}

\newcommand{\MailDaemon}{{\bf\ttfamily\color{nethack_blue}\sethlcolor{black}\hl{\&}}}

\newcommand{\Kobold}{{\bf\ttfamily\color{nethack_yellow}\sethlcolor{black}\hl{k}}}

\newcommand{\GnomeLord}{{\bf\ttfamily\color{nethack_blue}\sethlcolor{black}\hl{G}}}

\newcommand{\GnomeKing}{{\bf\ttfamily\color{nethack_magenta}\sethlcolor{black}\hl{G}}}

\newcommand{\BlackPudding}{{\bf\ttfamily\color{nethack_gray}\sethlcolor{black}\hl{P}}}

\newcommand{\Human}{{\bf\ttfamily\color{nethack_white}\sethlcolor{black}\hl{@}}}

\newcommand{\Oracle}{{\bf\ttfamily\color{nethack_light_blue}\sethlcolor{black}\hl{@}}}

\newcommand{\Medusa}{{\bf\ttfamily\color{nethack_light_green}\sethlcolor{black}\hl{@}}}

\newcommand{\Kraken}{{\bf\ttfamily\color{nethack_red}\sethlcolor{black}\hl{;}}}

\newcommand{\Chameleon}{{\bf\ttfamily\color{nethack_yellow}\sethlcolor{black}\hl{:}}}

\newcommand{\Gold}{{\bf\ttfamily\color{nethack_light_yellow}\sethlcolor{black}\hl{\$}}}
\newcommand{\StaircaseDown}{{\bf\ttfamily\color{nethack_light_gray}\sethlcolor{black}\hl{>}}}

\newcommand{\Food}{{\bf\ttfamily\color{nethack_yellow}\sethlcolor{black}\hl{\%}}}
\newcommand{\Apple}{{\bf\ttfamily\color{nethack_green}\sethlcolor{black}\hl{\%}}}
\newcommand{\Orange}{{\bf\ttfamily\color{nethack_orange}\sethlcolor{black}\hl{\%}}}
\newcommand{\Door}{{\bf\ttfamily\color{nethack_yellow}\sethlcolor{black}\hl{+}}}
\newcommand{\Boulder}{{\bf\ttfamily\color{nethack_light_gray}\sethlcolor{black}\hl{O}}}

\newcommand{\Weapon}{{\bf\ttfamily\color{nethack_cyan}\sethlcolor{black}\hl{)}}}
\newcommand{\Armor}{{\bf\ttfamily\color{nethack_cyan}\sethlcolor{black}\hl{[}}}
\newcommand{\Tile}{{\bf\ttfamily\color{nethack_white}\sethlcolor{black}\hl{.}}}
\newcommand{\HiddenTile}{{\bf\ttfamily\color{nethack_light_gray}\sethlcolor{black}\hl{.}}}

\newcommand{\Ice}{{\bf\ttfamily\color{nethack_light_cyan}\sethlcolor{black}\hl{.}}}
\newcommand{\Water}{{\bf\ttfamily\color{nethack_blue}\sethlcolor{black}\hl{\}}}}
\newcommand{\Passage}{{\bf\ttfamily\color{nethack_light_gray}\sethlcolor{black}\hl{\#}}}
\newcommand{\Trap}{{\bf\ttfamily\color{nethack_magenta}\sethlcolor{black}\hl{\textasciicircum}}}
\newcommand{\Spellbook}{{\bf\ttfamily\color{nethack_light_magenta}\sethlcolor{black}\hl{+}}}
\newcommand{\Scroll}{{\bf\ttfamily\color{nethack_white}\sethlcolor{black}\hl{?}}}
\newcommand{\Boots}{{\bf\ttfamily\color{nethack_yellow}\sethlcolor{black}\hl{[}}}
\newcommand{\Tool}{{\bf\ttfamily\color{nethack_cyan}\sethlcolor{black}\hl{(}}}
\newcommand{\Sack}{{\bf\ttfamily\color{nethack_yellow}\sethlcolor{black}\hl{(}}}
\newcommand{\Mirror}{{\bf\ttfamily\color{nethack_yellow}\sethlcolor{black}\hl{(}}}
\newcommand{\Towel}{{\bf\ttfamily\color{nethack_magenta}\sethlcolor{black}\hl{(}}}

\usepackage{hyperref}

\title{The NetHack Learning Environment}

\newcommand{\fair}{\texttt{\large +}}
\newcommand{\oxford}{\texttt{\large =}}
\newcommand{\nyu}{\texttt{\large *}}
\newcommand{\imperial}{\texttt{\large \#}}
\newcommand{\ucl}{\texttt{\large !}}

\author{%
    \textbf{Heinrich K{\"u}ttler}$^\fair{}$
    \textbf{Nantas Nardelli}$^\oxford{}$
    \textbf{Alexander H. Miller}$^\fair{}$\\
    \textbf{Roberta Raileanu}$^\nyu{}$
    \textbf{Marco Selvatici}$^\imperial{}$
    \textbf{Edward Grefenstette}$^{\fair{} \ucl{}}$
    \textbf{Tim Rockt{\"a}schel}$^{\fair{} \ucl{}}$\\[1em]
    $^\fair{}$Facebook AI Research ${}^\oxford{}$University of Oxford ${}^\nyu{}$New York University\\
    ${}^\imperial{}$Imperial College London ${}^\ucl{}$University College London\\[0.5em]
    \texttt{\{hnr,rockt\}@fb.com}
}

\begin{document}

\maketitle

\begin{abstract}
Progress in Reinforcement Learning (RL) algorithms goes hand-in-hand
with the development of challenging environments that test the limits
of current methods. While existing RL environments are either
sufficiently complex or based on fast simulation, they are rarely
both. Here, we present the~\nethackenv{} (\NLE{}), a scalable,
procedurally generated, stochastic, rich, and challenging environment
for RL research based on the popular single-player terminal-based
roguelike game, \nethack{}. We argue that NetHack is sufficiently
complex to drive long-term research on problems such as exploration,
planning, skill acquisition, and language-conditioned RL, while
dramatically reducing the computational resources required to gather a
large amount of experience. We compare \NLE{} and its task suite to
existing alternatives, and discuss why it is an ideal medium for
testing the robustness and systematic generalization of RL agents. We
demonstrate empirical success for early stages of the game using a
distributed Deep RL baseline and Random Network Distillation
exploration, alongside qualitative analysis of various agents trained
in the environment.
\NLE{} is open source and available
at \url{https://github.com/facebookresearch/nle}.
\end{abstract}

\section{Introduction}
Recent advances in (Deep) Reinforcement Learning (RL) have been driven
by the development of novel simulation environments, such as the
Arcade Learning Environment (ALE)~\citep{DBLP:journals/jair/BellemareNVB13},
StarCraft~\citep{synnaeve2016torchcraft, vinyals2017starcraft},
BabyAI~\citep{chevalier2018babyai}, Obstacle
Tower~\citep{DBLP:conf/ijcai/JulianiKBHTHCTL19}, Minecraft~\citep{DBLP:conf/ijcai/JohnsonHHB16, guss2019minerlcomp, DBLP:journals/corr/abs-1905-01978}, and Procgen Benchmark~\citep{cobbe2019procgen}.  These environments
introduced new challenges for state-of-the-art methods and
demonstrated failure modes of existing RL approaches.
For example, \emph{Montezuma's Revenge} highlighted that methods performing well on other ALE tasks were not able to successfully learn in this sparse-reward environment.
This sparked a long line of research on
novel methods for exploration~\citep[e.g.,][]{bellemare2016unifying,
  tang2017exploration, ostrovski2017count} and learning from
demonstrations~\citep[e.g.,][]{Hester2017DeepQF,
  DBLP:journals/corr/abs-1812-03381,DBLP:conf/nips/AytarPBP0F18}.
However, this progress has limits: the current best approach on this
environment, Go-Explore~\citep{ecoffet2019go,DBLP:journals/corr/abs-2004-12919}, overfits to specific
properties of ALE and Montezuma's Revenge.
While Go-Explore is an impressive solution for Montezuma's Revenge, it exploits
the determinism of environment transitions, allowing it to memorize sequences of actions that lead to previously visited states from which the agent can continue to explore.

We are interested in surpassing the limits of deterministic or repetitive settings and seek a simulation environment that is complex and modular enough to test various open research
challenges such as exploration, planning, skill acquisition, memory, and transfer. However, since state-of-the-art RL approaches still require millions or even billions of samples, simulation
environments need to be fast to allow RL agents to perform many interactions per second.
Among attempts to surpass the limits of deterministic or repetitive
settings, \emph{procedurally generated environments} are a promising
path towards testing systematic generalization of RL
methods~\citep[e.g.,][]{justesen2018illuminating,DBLP:conf/ijcai/JulianiKBHTHCTL19,DBLP:journals/corr/abs-1911-13071,cobbe2019procgen}. Here,
the game state is generated programmatically in every
episode, making it extremely unlikely for an agent to visit the exact
state more than once during its lifetime.  Existing procedurally generated RL environments
are either costly to
run~\citep[e.g.,][]{vinyals2017starcraft,DBLP:conf/ijcai/JohnsonHHB16,DBLP:conf/ijcai/JulianiKBHTHCTL19}
or are, as we argue, of limited complexity~\citep[e.g.,][]{gym_minigrid,DBLP:conf/icml/CobbeKHKS19,DBLP:journals/corr/BeattieLTWWKLGV16}.

To address these issues, we present the \nethackenv{} (\NLE{}), a procedurally generated environment that strikes a balance between complexity and speed. It is a fully-featured
\emph{Gym} environment~\citep{DBLP:journals/corr/BrockmanCPSSTZ16} around the popular open-source terminal-based
single-player turn-based ``dungeon-crawler''
game, \nethack{}~\citep{NetHackOrg}.  Aside from procedurally generated content, \nethack{} is an
attractive research platform as it contains hundreds of enemy and
object types, it has complex and stochastic environment dynamics, and
there is a clearly defined goal (descend the dungeon, retrieve an
amulet, and ascend).  Furthermore, \nethack{} is difficult to master
for human players, who often rely on external knowledge to learn about strategies and NetHack's complex
dynamics and secrets.\footnote{``NetHack is largely based on discovering secrets
  and tricks during gameplay. It can take years for one to become
  well-versed in them, and even experienced players routinely discover
  new ones.'' \citep{elitmod}} Thus, in addition to a guide
book~\cite{raymond1987guide, raymond2020guide} released with \nethack{} itself, many extensive community-created documents exist, outlining various strategies for the game~\citep[e.g.,][]{nhwiki,NetHackSpoilers}.

In summary, we make the following core contributions:
\begin{enumerate*}[label=(\roman*)]
\item we present \NLE{}, a fast but complex and feature-rich Gym
  environment for RL research built around the popular terminal-based
  game, \nethack{},
\item we release an initial suite of tasks in the environment
  and demonstrate that novel tasks can be added easily,
\item we introduce baseline models trained using IMPALA~\citep{DBLP:conf/icml/EspeholtSMSMWDF18} and Random Network Distillation (RND) \citep{DBLP:conf/iclr/BurdaESK19}, a popular exploration bonus, resulting in agents that learn diverse policies for early stages of \nethack{}, and
\item we demonstrate the benefit of \nethack{}'s symbolic observation space by presenting in-depth qualitative analyses of
  trained agents.
\end{enumerate*}

\section{NetHack: a Frontier for Reinforcement Learning Research}
In traditional so-called \emph{roguelike} games (e.g., \rogue{}, \hack{}, \nethack{},
and \dcss{}) the player acts turn-by-turn in a procedurally generated
grid-world environment, with game dynamics strongly focused on
exploration, resource management, and continuous discovery of entities
and game mechanics~\citepalias{berlin}.
These games are designed to provide a steep learning curve and a
constant level of challenge and surprise to the player.  They are
generally extremely difficult to win even once, let alone to master, i.e., win regularly and multiple times in a row.

As advocated by
\cite{justesen2018illuminating,DBLP:conf/ijcai/JulianiKBHTHCTL19,cobbe2019procgen},
procedurally generated environments are a promising direction for testing
systematic generalization of RL agents.  We argue that such environments need to be both sufficiently complex and fast to run to serve as a
challenging long-term research testbed.
In \Cref{sec:nethack}, we illustrate that \nethack{}
contains many desirable properties, making it an excellent candidate for driving long-term research in RL.
We introduce \NLE{} in \Cref{sec:nle}, an initial
suite of tasks in \Cref{sec:tasks}, an evaluation protocol for measuring progress towards \emph{solving} NetHack in \Cref{sec:eval}, as well as baseline models in \Cref{sec:models}.

\subsection{NetHack}
\label{sec:nethack}
NetHack is one of the oldest and most popular roguelikes,
originally released in 1987 as a successor to \emph{Hack}, an
open-source implementation of the original \emph{Rogue} game.
At the beginning of the game, the player takes the role of a hero who is placed
into a dungeon and tasked with finding the \emph{Amulet of Yendor} to offer it to an in-game deity. To do so, the player
has to descend to the bottom of over 50 procedurally generated
levels to retrieve the amulet and then subsequently escape the
dungeon, unlocking five extremely challenging final levels (the four Elemental Planes and the Astral Plane).

\begin{figure*}[t!]
    \centering
    \includegraphics[trim=0 10 10 3, clip, width=\textwidth]{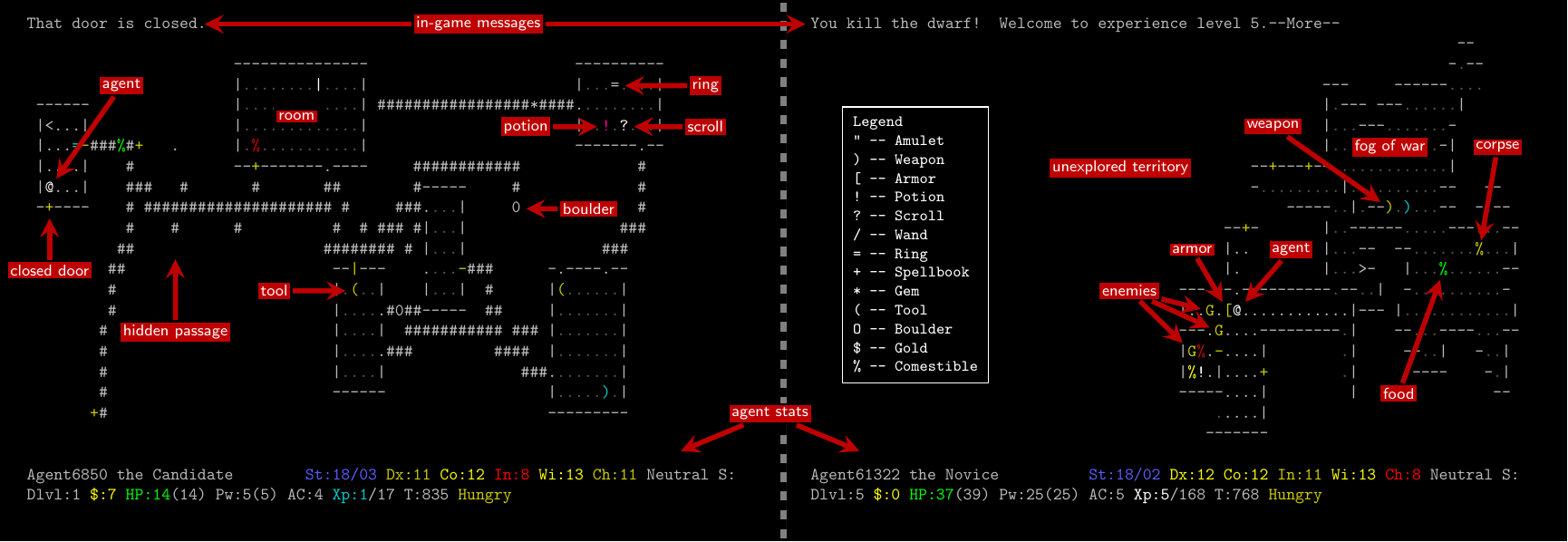}
    \caption{Annotated example of an agent at two different stages in
      \nethack{} (Left: a procedurally generated first level of the
      Dungeons of Doom, right: Gnomish Mines). A larger version of this figure is displayed in Figure~\ref{fig:levelBig} in the appendix.}
    \label{fig:level}
\end{figure*}

Many aspects of the game are procedurally generated and follow
stochastic dynamics.  For example, the overall structure of the
dungeon is somewhat linear, but the exact location of places of
interest (e.g., the \emph{Oracle}) and the structure of branching
sub-dungeons (e.g., the \emph{Gnomish Mines}) are determined
randomly.
The procedurally generated content of each level makes it highly
unlikely that a player will ever experience the exact same situation more than once.
This provides a fundamental challenge to learning systems and a degree
of complexity that enables us to more effectively evaluate an agent's
ability to generalize.
It also disqualifies current state-of-the-art exploration methods such as Go-Explore~\citep{ecoffet2019go,DBLP:journals/corr/abs-2004-12919} that are based on a goal-conditioned policy to navigate to previously visited states.
Moreover, states in NetHack are composed of hundreds of possible symbols,
resulting in an enormous combinatorial observation space.\footnote{Information about the over 450 items and
580 monster types, as well as environment dynamics
involving these entities can be found in the NetHack
Wiki~\citep{nhwiki} and to some extent in the NetHack Guidebook \citep{raymond2020guide}.}
It is an open question how to best project this symbolic space to a low-dimensional representation appropriate for methods like Go-Explore.
For example, \citeauthor{ecoffet2019go}'s heuristic of downsampling images of states to measure their similarity to be used as an exploration bonus will likely not work for large symbolic and procedurally generated environments.
NetHack provides further variation by different hero roles (e.g., monk, valkyrie, wizard, tourist), races (human, elf, dwarf, gnome, orc) and random starting inventories (see \Cref{sec:moreaboutnethack} for details).
Consequently, NetHack poses unique challenges to the research community and requires novel ways to determine state similarity and, likely, entirely new exploration frameworks.

\begin{wrapfigure}{r}{0.5\textwidth}
  \begin{center}
    \vspace{-3ex}
    \includegraphics[trim=5 15 10 2, clip, width=0.5\textwidth]{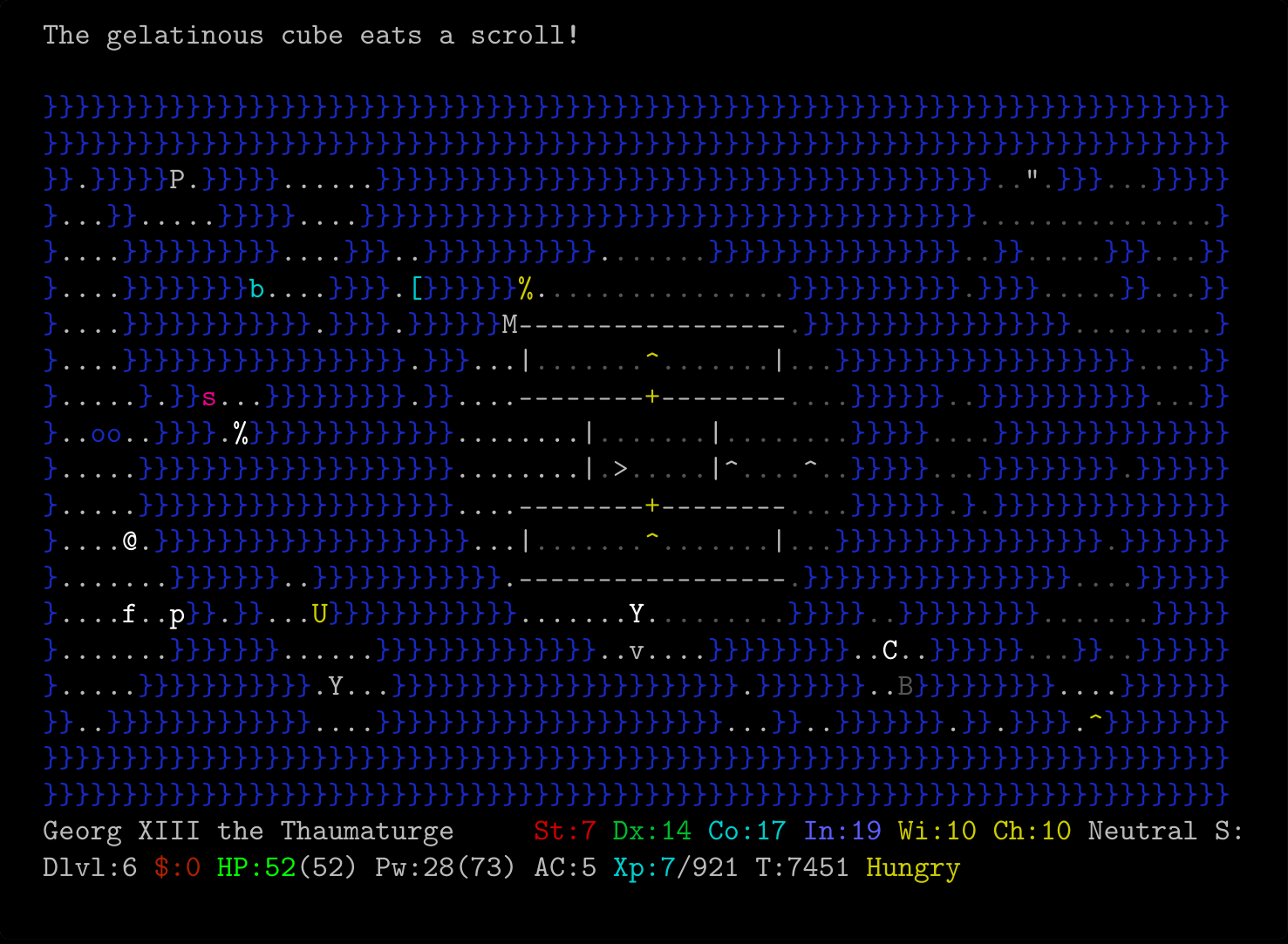}
  \end{center}
  \caption{The hero (\protect\Human{}) has to cross water
  (\protect\Water{}) to get past Medusa (\protect\Medusa{}, out of the
  hero's line of sight) down the staircase (\protect\StaircaseDown{}) to the
  next level.}
  \label{fig:medusa}
\end{wrapfigure}
To provide a glimpse into the complexity of \nethack{}'s environment
dynamics, we closely follow the educational example given by ``Mr Wendal'' on
YouTube.\footnote{\href{https://youtube.com/watch?v=SjuTyJlgLJ8}{\nolinkurl{youtube.com/watch?v=SjuTyJlgLJ8}}}
At a specific point in the game, the hero has to get past \emph{Medusa's Island} (see \Cref{fig:medusa} for an example).
Medusa's Island is surrounded by water \Water{} that the agent has to cross.
Water can rust and
corrode the hero's metallic weapons \Weapon{} and armor \Armor{}.
Applying a can of grease~\Tool{} prevents rusting and corrosion.  Furthermore,
going into water will make a hero's inventory wet, erasing scrolls
\Scroll{} and spellbooks \Spellbook{} that they carry.  Applying a can
of grease to a bag or sack~\Sack{} will make it a waterproof container for
items.  But the sea can also contain a kraken \Kraken{} that can
grab and drown the hero, leading to instant death.  Applying a can of
grease to a hero's armor prevents the kraken from grabbing the hero.
However, a cursed can of grease will grease the hero's hands
instead and they will drop their weapon and rings.  One can use a
towel~\Towel{} to wipe off grease.
To reach Medusa \Medusa{}, the hero can alternatively use magic to freeze the water and turn it into walkable ice~\Ice{}. Wearing snow boots \Boots{} will help the hero not to slip.
When Medusa is in the hero's line of sight, her gaze will petrify and instantly kill---the hero should use a towel to cover their eyes to fight Medusa, or even apply a mirror~\Mirror{} to petrify her with her own gaze.

There are many other entities a hero must learn to face, many of which appear rarely even across multiple games, especially the most powerful monsters.
These entities
are often compositional, for example a monster might be a wolf
\Wolf{}, which shares some characteristics with other in-game canines
such as coyotes \Coyote{} or hell hounds \HellHound{}.
To help a player learn, NetHack provides in-game messages describing many of the
hero's interactions (see the top of \Cref{fig:level}).\footnote{An example interaction after applying a figurine of an Archon: ``{\it You set the figurine on the ground and it transforms. You get a bad feeling about this. The Archon hits! You are blinded by the Archon's radiance! You stagger\ldots\ It hits! You die\ldots\ But wait\ldots\ Your medallion feels warm! You feel much better! The medallion crumbles to dust! You survived that attempt on your life.}''}
Learning to capture these interesting and somewhat realistic albeit abstract dynamics poses challenges for multi-modal and language-conditioned RL \cite{luketina2019survey}.

NetHack is an extremely long game. Successful expert episodes usually last
tens of thousands of turns, while average successful runs can easily
last hundreds of thousands of turns, spawning multiple days of
play-time. Compared to testbeds with long episode horizons such as
StarCraft and Dota 2, NetHack's ``episodes'' are one or two orders of
magnitude longer, and they wildly vary depending on the policy.
Moreover, several official \emph{conducts} exist in NetHack that make the game even more challenging, e.g., by not wearing any armor throughout the game (see \Cref{sec:moreaboutnethack} for more).

Finally, in comparison to other classic roguelike games, NetHack's
popularity has attracted a larger number of contributors to its
community.  Consequently, there exists a comprehensive game wiki~\citep{nhwiki} and many so-called spoilers~\citep{NetHackSpoilers} that provide advice to players. Due to the randomized nature of NetHack, this advice is general in nature  (e.g., explaining the behavior of various entities) and not a step-by-step guide.
These texts could be used for language-assisted RL along the lines of \citep{Zhong2019RTFMGT}. Lastly, there is also a large public repository of human replay data (over five million games) hosted on the
NetHack Alt.org (NAO)
servers, with hundreds of finished games per day on average \citep{AltOrgNethack}.
This extensive
dataset could spur research advances in imitation learning, inverse
RL, and learning from demonstrations
\citep{Abbeel2004ApprenticeshipLV, Argall2009ASO}.

\subsection{The NetHack Learning Environment}
\label{sec:nle}

The \nethackenv{} (\NLE{}) is built on NetHack~3.6.6, the 36th public
release of NetHack, which was released on March 8th, 2020 and is the
latest available version of the game at the time of publication of
this paper.
\NLE{} is designed to provide a common,
turn-based (i.e., synchronous) RL interface around
the standard terminal interface of NetHack.  We use the game
\emph{as-is} as the backend for our \NLE{} environment, leaving the game dynamics unchanged. We added to the source code more
control over the random number generator for seeding the environment,
as well as various modifications to expose the game's internal state
to our Python frontend.

By default, the observation space consists of the elements
\textit{glyphs}, \textit{chars}, \textit{colors}, \textit{specials},
\textit{blstats}, \textit{message}, \textit{inv\_glyphs}, \textit{inv\_strs},
\textit{inv\_letters}, as well as \textit{inv\_oclasses}.
The elements $\textit{glyphs}$, \textit{chars}, \textit{colors},
and \textit{specials} are tensors representing
the (batched) 2D symbolic observation of the dungeon; $\textit{blstats}$ is a vector
of agent coordinates and other character attributes (``bottom-line
stats'', e.g., health
points, strength, dexterity, hunger level; normally displayed in the
bottom area of the GUI), \textit{message} is a tensor representing
the current message shown to the player (normally displayed in the top
area of the GUI), and the \textit{inv\_*} elements are padded tensors
representing the hero's inventory items. More details about the
default observation space and possible extensions can be found
in \Cref{sec:observation}.

The environment has 93 available actions, corresponding to all the
actions a human player can take in \nethack{}. More precisely, the
action space is composed of 77 command actions and 16 movement
actions. The movement actions are split into eight ``one-step'' compass
directions (i.e., the agent moves a single step in a given direction)
and eight ``move far'' compass directions (i.e., the agent moves in the
specified direction until it runs into some entity). The 77 command
actions include \emph{eating}, \emph{opening}, \emph{kicking},
\emph{reading}, \emph{praying} as well as many others. We refer the
reader to~\Cref{sec:actions} as well as to the NetHack Guidebook \cite{raymond2020guide}
for the full table of actions and NetHack commands.

\NLE{} comes with a Gym
interface~\citep{DBLP:journals/corr/BrockmanCPSSTZ16} and includes
multiple pre-defined tasks with different reward functions and action
spaces (see next section and \Cref{sec:taskdetails} for details). We designed the interface to be
lightweight, achieving competitive speeds with Gym-based ALE (see
\Cref{sec:envsspeed} for a rough comparison).
Finally, \NLE{} also includes a dashboard to analyze NetHack runs
recorded as terminal \texttt{tty} recordings. This allows \NLE{} users
to analyze replays of the agent's behavior at an arbitrary speed and
provides an interface to visualize action distributions and game
events (see \Cref{sec:dashboard} for details).
\NLE{} is available under an open source license at \url{https://github.com/facebookresearch/nle}.

\subsection{Tasks}
\label{sec:tasks}

\NLE{} aims to make it easy for researchers to probe the
behavior of their agents by defining new tasks with only a few lines
of code, enabled by \nethack{}'s symbolic observation space as well as
its rich entities and environment dynamics. To demonstrate that
\nethack{} is a suitable testbed for advancing RL, we release a set of
initial tasks for tractable subgoals in the game: navigating to a \textbf{staircase} down to the next level, navigating to a staircase while being accompanied by a \textbf{pet}, locating and \textbf{eat}ing edibles, collecting \textbf{gold}, maximizing in-game \textbf{score}, \textbf{scout}ing to discover unseen parts of the dungeon, and finding the \textbf{oracle}. These tasks are described in detail in \Cref{sec:taskdetails}, and, as we demonstrate in our experiments, lead to unique challenges and diverse behaviors of trained agents.

\subsection{Evaluation Protocol}
\label{sec:eval}
We lay out a protocol and provide guidance for evaluating future work on \NLE{} in a reproducible manner.
The overall goal of \NLE{} is to train agents that can solve \nethack{}.
An episode in the full game of \nethack{} is considered solved if the agent retrieves the \emph{Amulet of Yendor} and offers it to its co-aligned deity in the Astral Plane, thereby ascending to demigodhood.
We declare \NLE{} to be solved once agents can be trained to consecutively ascend (ten episodes without retry) to demigodhood on unseen seeds given a random role, race, alignment, and gender combination.
Since the environment is procedurally generated and stochastic, evaluating on held-out unseen seeds ensures we test systematic generalization of agents.
As of October 2020, NAO reports the longest \textit{streak} of human ascensions on NetHack~3.6.$x$ to be 61; the role, race, etc. are not necessarily randomized for these ascension streaks.
Since we believe that this goal is out of reach for machine learning approaches in the foreseeable future, we recommend comparing models on the score task in the meantime.
Using \nethack{}'s in-game score as the measure for progress has caveats.
For example, expert human players can solve \nethack{} while minimizing the score \citep[see][``Score'' entry, for details]{nhwiki}. NAO reports ascension scores for \nethack~3.6.$x$ ranging from the low hundreds of thousands to tens of millions.
Although we believe training agents to maximize the in-game score is likely insufficient for solving the game, the in-game score is still a sensible proxy for incremental progress on \NLE{} as it is a function of, among other things, the dungeon depth that the agent reached, the number of  enemies it killed, the amount of gold it collected, as well as the knowledge it gathered about potions, scrolls, and wands.

When reporting results on \NLE{}, we require future work to state the full character specification~(e.g.,~\verb~mon-hum-neu-mal~), all \nethack{} options that were used (e.g., whether or not \emph{autopickup} was used), which actions were allowed (see \Cref{table:commandactions}), which actions or action-sequences were hard-coded~(e.g.,~engraving \citep[see][``Elbereth'' as an example]{nhwiki}) and how many different seeds were used during training. We ask to report the average score obtained on $1000$ episodes of randomly sampled and previously unseen seeds.
We do not impose any restrictions during training, but at test time any save scumming (i.e., saving and loading previous checkpoints of the episode) or manipulation of the random number generator \citep[e.g.,][]{swagginzzz} is forbidden.

\subsection{Baseline Models}
\label{sec:models}

\begin{figure}[t]
\centering
\begin{minipage}{.53\textwidth}
  \centering
  \includegraphics[width=1.0\linewidth]{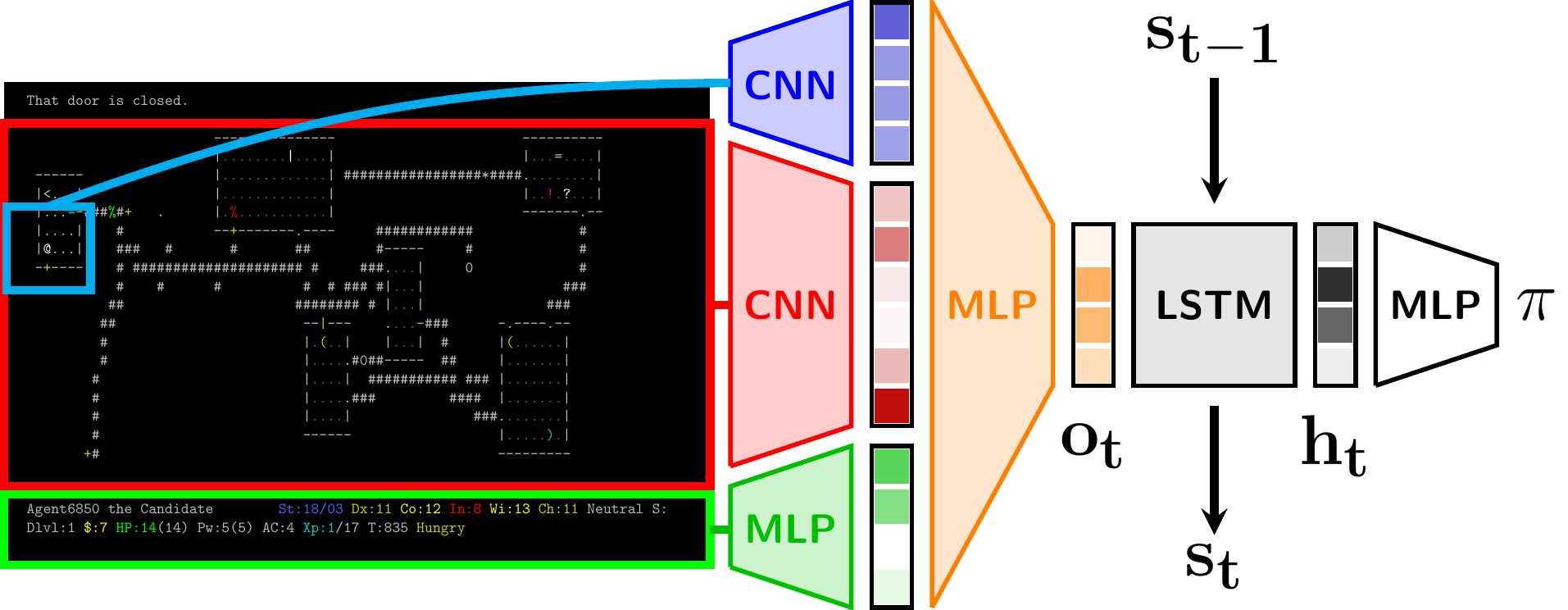}
  \captionof{figure}{Overview of the core architecture of the baseline models released with \NLE{}. A larger version of this figure is displayed in Figure~\ref{fig:modelBig} in the appendix.}
  \label{fig:model}
\end{minipage}\hfill
\begin{minipage}{.44\textwidth}
  \centering
  \includegraphics[width=1.0\linewidth]{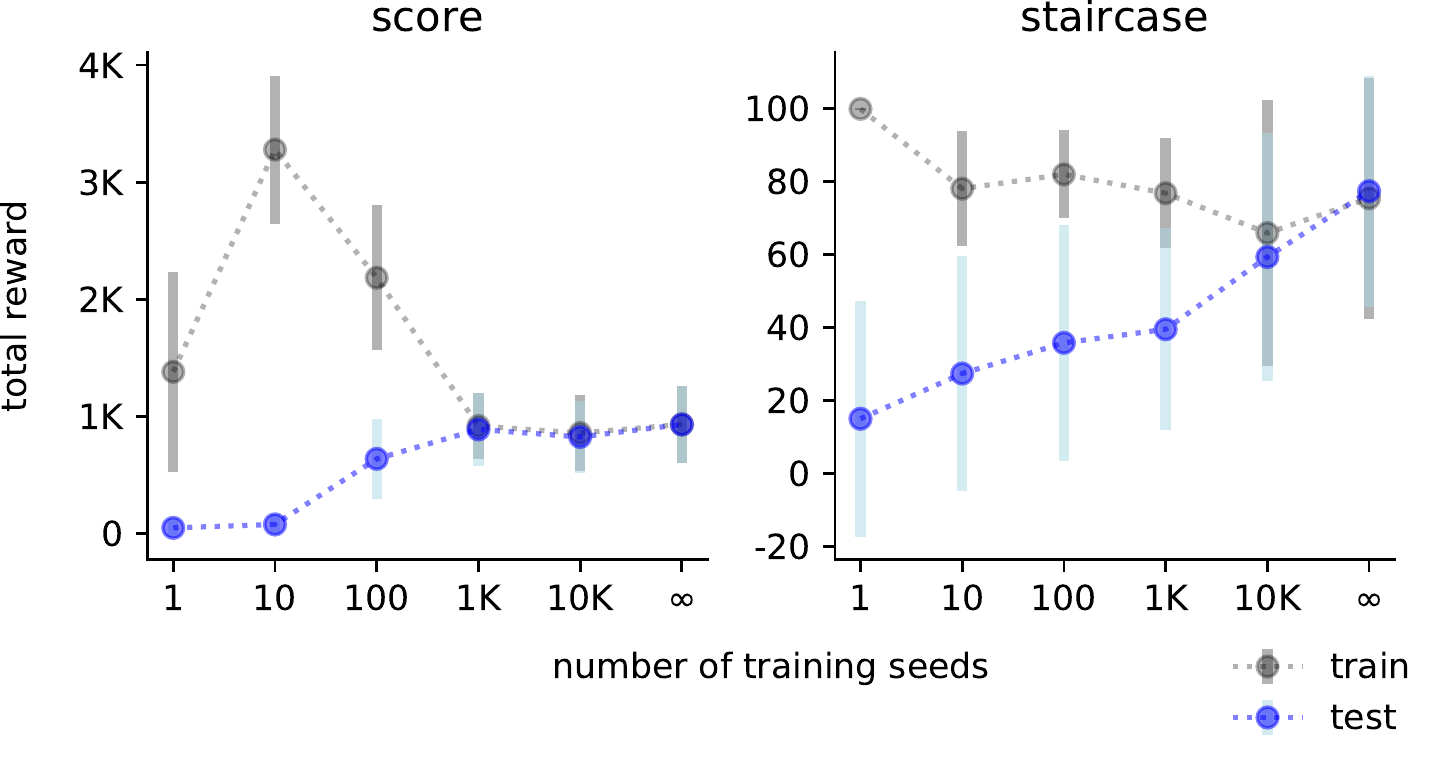}
  \captionof{figure}{Training and test performance when training on restricted sets of seeds.}
  \label{fig:results_seeds}
\end{minipage}
\end{figure}

For our baseline models, we encode the multi-modal observation $o_t$ as
follows.  Let the observation $o_t$ at time step $t$ be a tuple $(g_t,
z_t)$ consisting of the $21\times 79$ matrix of glyph identifiers and a
$21$-dimensional vector containing agent stats such as its
$(x,y)$-coordinate, health points, experience level, and so on.  We
produce three dense representations based on the observation (see
\Cref{fig:model}).  For every of the $5991$ possible glyphs in
\nethack{} (monsters, items, dungeon features, etc.), we learn a
$k$-dimensional vector embedding.  We apply a
ConvNet (red) to all visible glyph embeddings as well as
another ConvNet (blue) to
the $9\times 9$ crop of glyphs around the agent to create a dedicated egocentric representation for improved
generalization~\citep{DBLP:journals/corr/abs-1910-00571,ye2020rotation}.
We found this egocentric representation to be an important component during preliminary experiments.
Furthermore, we use an MLP to encode the hero's stats (green).  These
vectors are concatenated and processed by another MLP to produce
a low-dimensional latent representation $\mathbf{o}_t$ of the
observation.  Finally, we employ a recurrent policy parameterized
by an LSTM~\citep{DBLP:journals/neco/HochreiterS97} to obtain the
action distribution.
For baseline results on the tasks above, we use a reduced action space that includes the movement, search, kick, and eat actions.

For the main experiments, we train the agent's policy for 1B steps in the environment using
IMPALA~\citep{DBLP:conf/icml/EspeholtSMSMWDF18} as implemented in
TorchBeast~\citep{Kttler2019TorchBeastAP}.  Throughout training, we
change \nethack{}'s seed for procedurally generating the environment
after every episode.
To demonstrate NetHack's variability based on the character configuration, we train with four different agent characters: a neutral human male monk (\verb~mon-hum-neu-mal~), a lawful dwarf female valkyrie (\verb~val-dwa-law-fem~), a chaotic elf male wizard (\verb~wiz-elf-cha-mal~), and a neutral human female tourist (\verb~tou-hum-neu-fem~).
More implementation details
can be found in \Cref{appendix:baseline}.

In addition, we present results using Random Network Distillation (RND)~\citep{DBLP:conf/iclr/BurdaESK19}, a popular exploration technique for Deep RL.
As previously discussed, exploration techniques which require returning to previously visited states such as Go-Explore are not suitable for use in \NLE{}, but RND does not have this restriction.
RND encourages agents to visit unfamiliar states by using the prediction error of a fixed random network as an intrinsic exploration reward, which has proven effective for hard exploration games such as Montezuma's Revenge~\citep{DBLP:conf/iclr/BurdaEPSDE19}.
The intrinsic reward obtained from RND can create ``reward bridges'' between states which provide sparse extrinsic environmental rewards, thereby enabling the agent to discover new sources of extrinsic reward that it otherwise would not have reached.
We replace the baseline network's pixel-based feature extractor with the symbolic feature extractor described above for the baseline model, and use the best configuration of other RND hyperparameters documented by the authors (see \Cref{appendix:rnd} for full details).

\section{Experiments and Results}
\label{sec:experiments}
We present quantitative results on the suite of tasks included in \NLE{} using a standard distributed Deep RL baseline and a popular exploration method, before additionally analyzing agent behavior qualitatively.
For each model and character combination, we present results of the mean episode return over the last $100$ episodes averaged for five runs in \Cref{fig:results}. We discuss results for
individual tasks below (see \Cref{tab:results:score} in the appendix for full details).

\paragraph{Staircase:}
Our agents learning to navigate the dungeon to the staircase \StaircaseDown{} with a success rate of $77.26\%$ for the monk, $50.42\%$ for the tourist, $74.62\%$ for the valkyrie, and $80.42\%$ for the wizard.
What surprised us is that agents learn to reliably kick in locked doors. This is a costly action to explore as the agent loses health points and might even die when
accidentally kicking against walls.
Similarly, the agent has to learn to reliably search for hidden passages and secret doors.
Often, this involves using the search action many times in a row, sometimes even at many locations on the map (e.g., around all walls inside a room).
Since \NLE{} is procedurally generated, during training agents might encounter easier environment instances and use the acquired skills to accelerate learning on the harder ones \cite{DBLP:journals/corr/abs-1911-13071,cobbe2019procgen}.
With a small probability, the staircase down might be generated  near the agent's starting position.
Using RND exploration, we observe substantial gains in the success rate for the monk ($+13.58$pp), tourist ($+6.52$pp) and valkyrie ($+16.34$pp) roles, while lower results for wizard roles ($-12.96$pp).

\paragraph{Pet:}
Finding the staircase while taking care of the hero's pet (e.g., the starting kitten~\Kitten{} or little dog~\LittleDog{}) is a harder task as the pet might get killed or fall into a
trap door, making it impossible for the agent to successfully complete the episode.
Compared to the staircase task, the agent success rates are generally lower ($62.02\%$ for monk, $25.66\%$ for tourist, $63.30\%$ for valkyrie, and wizard $66.80\%$). Again, RND exploration provides consistent and substantial gains.

\paragraph{Eat:}
This tasks highlights the importance of testing with different
character classes in \nethack{}.  The monk and tourist start with a number edible items (e.g., food rations~\Food{}, apples~\Apple{} and oranges~\Orange{}).  A sub-optimal strategy is to consume all of these
comestibles right at the start of the episode, potentially risking choking to death.  In contrast, the other roles have to hunt for food, which our agents learn to do slowly over time for the valkyrie and wizard roles.
By having more pressure to quickly learn a sustainable
food strategy, the valkyrie learns to outlast other roles and survives the longest in the game (on average $1713$ time steps).
Interestingly, RND exploration leads to consistently worse results for this task.

\paragraph{Gold:}
Locating gold \Gold{} in \nethack{} provides a relatively sparse reward signal. Still, our agents learn to collect decent amounts during training and learn to descend to deeper dungeon levels in search for more. For example, monk agents reach dungeon level $4.2$ on average for the CNN baseline and even $5.0$ using RND exploration.

\paragraph{Score:}
As discussed in \Cref{sec:eval}, we believe this task is the best candidate for comparing future methods regarding progress on \nethack{}. However, it is questionable whether a reward function based on \nethack{}'s in-game score is sufficient for training agents to solve the game. Our agents average at a score of $748$ for monk, $11$ for tourist, $573$ for valkyrie, and $314$ for wizard, with RND exploration again providing substantial gains (e.g. increasing the average score to $780$ for monk).
The resulting agents explore much of the early stages of the game, reaching dungeon level $5.4$ on average for the monk with the deepest descent to level $11$ achieving a high score of $4260$ while leveling up to experience level $7$ (see \Cref{tab:results:top} in the appendix).

\paragraph{Scout:}
The scout task shows a trend that is similar to the score task.
Interestingly, we observe a lower experience level and in-game score, but agents descend, on average,
similarly deep into the dungeon (e.g. level $5.5$ for monk).
This is sensible, since a policy that avoids to
fight monsters, thereby lowering the chances of premature death, will not
increase the in-game score as fast or level up the character as quickly, thus keeping the difficulty of spawned monsters low.
We note that delaying to level up in order to avoid
encountering stronger enemies early in the game is a known strategy human
players adopt in \nethack{} \citep[e.g.][``Why do I keep dying?''
entry, January 2019 version]{nhwiki}.

\paragraph{Oracle:}
None of our agents find the Oracle \Oracle{} (except for one lucky valkyrie episode). Locating the Oracle is a difficult exploration task.
Even if the agent learns to make its way down the dungeon levels, it needs to search many, potentially branching, levels of the dungeon.
Thus, we believe this task serves as a challenging benchmark for exploration methods in procedurally generated environments in the short term.
Long term, many tasks harder than this (e.g.,
reaching \emph{Minetown}, \emph{Mines' End}, \emph{Medusa's Island}, \emph{The Castle}, \emph{Vlad's Tower}, \emph{Moloch's Sanctum} etc.) can be easily defined in \NLE{} with very few lines of code.

\begin{figure*}[t!]
    \centering \includegraphics[width=1.0\textwidth]{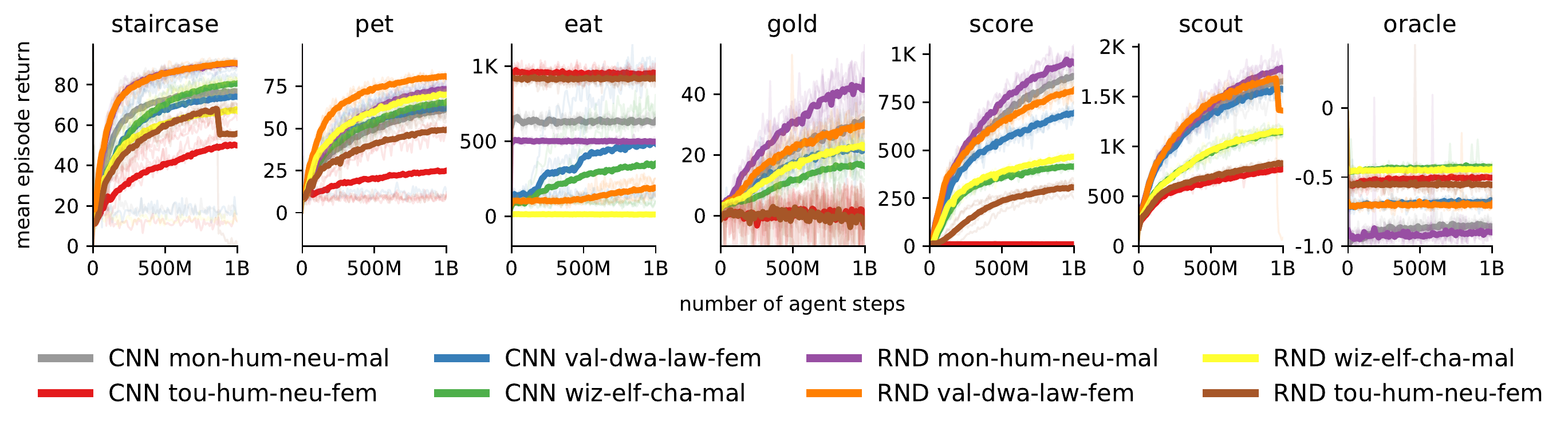}
    \caption{Mean return of the last $100$ episodes averaged
      over five runs.}
    \label{fig:results}
\end{figure*}

\subsection{Generalization Analysis}
Akin to \citep{cobbe2019procgen}, we evaluate agents trained on a limited set of seeds while still testing on 100 held-out seeds.
We find that test performance increases monotonically with the size of the set of seeds that the agent is trained on. \Cref{fig:results_seeds} shows this effect for the score and staircase tasks. Training only on a limited number of seeds leads to high training performance, but poor generalization.
The gap between training and test performance becomes narrow when training with at least $1000$ seeds, indicating that at that point agents are exposed to sufficient variation during training to make memorization infeasible.
We also investigate how model capacity affects performance by comparing agents with five different hidden sizes for the final layer (of the architecture described in \Cref{sec:models}).
\Cref{fig:results_size} in the appendix shows that increasing the model capacity improves results on the score but not on the staircase task, indicating that it is an important hyperparameter to consider, as also noted by~\citep{cobbe2019procgen}.

\subsection{Qualitative Analysis}
We analyse the cause for death of our agents during training and present results in \Cref{fig:death_analysis} in the appendix.
We notice that starvation and traps become a less prominent cause of death over time, most likely because our agents, when starting to learn to descend dungeon levels and fight monsters, are more likely to die in combat before they starve or get killed by a trap.
In the score and scout tasks, our agents quickly learn to avoid eating rotten corpses, but food poisoning becomes again prominent towards the end of training.

We can see that gnome lords \GnomeLord{}, gnome kings \GnomeKing{}, chameleons \Chameleon{}, and even mind flayers \MindFlayer{} become a
more prominent cause of death over time, which can be explained with our agents leveling up and descending deeper into the dungeon.
Chameleons are a particularly interesting entity in NetHack as they regularly change their form to a random animal or monster, thereby adversarially confusing our agent with rarely seen symbols for which it has not yet learned a meaningful representation  (similar to unknown words in natural language processing).
We release a set of high-score recordings of our agents (see \Cref{sec:runsamples} on how to view them via a browser or terminal).

\section{Related Work}

Progress in RL has historically been achieved both by algorithmic
innovations as well as development of novel environments to train and
evaluate agents. Below, we review recent RL environments and delineate their strengths and weaknesses as testbeds for current methods and future
research.

\paragraph{Recent Game-Based Environments:}
Retro video games have been a major catalyst for Deep RL research. ALE~\citep{DBLP:journals/jair/BellemareNVB13} provides a
unified interface to Atari 2600 games, which enables testing of RL algorithms
on high-dimensional visual observations quickly and cheaply,
resulting in numerous Deep RL publications over the years \citep{arulkumaran2017deep}. The
\emph{Gym Retro} environment \citep{nichol2018gotta} expands the list
of classic games, but focuses on
evaluating visual generalization and transfer learning on a single
game, \emph{Sonic The Hedgehog}.

Both \emph{StarCraft: BroodWar} and \emph{StarCraft II} have been
successfully employed as RL environments~\citep{synnaeve2016torchcraft, vinyals2017starcraft} for
research on, for example, planning~\citep{ontanon2013survey,nardelli2018value}, multi-agent systems~\citep{foerster2017stabilising, samvelyan2019starcraft}, imitation
learning~\citep{vinyals2019grandmaster}, and model-free reinforcement
learning~\citep{vinyals2019grandmaster}. However,
the complexity of these games creates a high entry barrier both in
terms of computational resources required as well as intricate baseline
models that require a high degree of domain knowledge to be extended.

3D games have proven to be useful testbeds for tasks such as
navigation and embodied
reasoning. \emph{Vizdoom}~\citep{kempka2016vizdoom} modifies the
classic first-person shooter game \emph{Doom} to construct an API for
visual control; \emph{DeepMind
  Lab}~\citep{DBLP:journals/corr/BeattieLTWWKLGV16} presents a game
engine based on \emph{Quake III Arena} to allow for the creation of
tasks based on the dynamics of the original game; \emph{Project
  Malmo}~\citep{DBLP:conf/ijcai/JohnsonHHB16}, MineRL~\citep{guss2019minerlcomp} and
\emph{CraftAssist}~\citep{DBLP:journals/corr/abs-1905-01978} provide
visual and symbolic interfaces to the popular \emph{Minecraft} game.
While Minecraft is also procedurally generated and has complex environment dynamics that an agent needs to learn about, it is much more computationally demanding than
\nethack{} (see \Cref{table:envscomparison} in the appendix). As a consequence, the focus has been on learning from demonstrations \citep{guss2019minerlcomp}.

More recent work has produced game-like environments with procedurally
generated elements, such as the \emph{Procgen Benchmark}~\cite{cobbe2019procgen},
\emph{MazeExplorer}~\citep{harrieslee2019}, and the \emph{Obstacle
  Tower} environment~\citep{DBLP:conf/ijcai/JulianiKBHTHCTL19}.  However, we
argue that, compared to \nethack{} or Minecraft, these environments do
not provide the depth likely necessary to serve as long-term RL
testbeds due to limited number of entities and environment interactions that agents have to learn to master.
In contrast, \nethack{} agents have to acquire knowledge about complex environment dynamics of hundreds of entities (dungeon features, items, monsters etc.) to do well in a game that humans often take years of practice to solve.

In conclusion, none of the current benchmarks combine a fast simulator
with a procedurally generated environment, a hard exploration problem,
a wide variety of complex environment dynamics, and numerous types of
static and interactive entities. The unique combination of challenges
present in NetHack makes \NLE{} well-suited for driving research
towards more general and robust RL algorithms.

\paragraph{Roguelikes as Reinforcement Learning Testbeds:}
We are not the first to argue for roguelike games to be used as
testbeds for RL.
\citet{asperti2019crawling} present an interface to Rogue, the very
first roguelike game and one of the simplest roguelikes in terms of
game dynamics and difficulty. They show that policies trained with
model-free RL algorithms can successfully learn rudimentary
navigation.
Similarly, \citet{kanagawa2019rogue} present an environment inspired
by Rogue that provides a parameterizable generation of Rogue levels.
Like us, \citet{dannenhauer2019dungeon} argue that roguelike games
could be a useful RL testbed. They discuss the roguelike game
\emph{Dungeon Crawl Stone Soup}, but their position paper
provides neither an RL environment nor experiments to validate their claims.

Most similar to our work is \emph{gym\_nethack}~\citep{campbell-17-learning, campbell2018exploration}, which offers a Gym environment based on \nethack~3.6.0. We commend the authors for introducing NetHack as an RL environment, and to the best of our knowledge they were the first to suggest the idea. However, there are several design choices that limit the impact and longevity of their version as a research testbed.
First, they heavily modified \nethack{} to enable agent interaction. In the process, \emph{gym\_nethack} disables various crucial game mechanics to simplify the game, its environment dynamics, and the resulting optimal policies.
This includes removing obstacles like boulders, traps, and locked doors as well as all item identification mechanics, making items much easier to employ and the overall environment much closer to its simpler predecessor, Rogue.
Additionally, these modifications tie the environment to a particular version of the game. This is not ideal as
(i) players tend to use new versions of the game as they are released, hence, publicly available human data becomes progressively incompatible, thereby limiting the amount of data that can be used for learning from demonstrations;
(ii) older versions of \nethack{} tend to include well-documented exploits which
may be discovered by agents (see \Cref{sec:bots} for exploits used in programmatic bots).
In contrast, \NLE\ is designed to make the interaction with \nethack{} as close as possible to the one experienced by humans playing the full game.
\NLE\ is the only environment exposing the entire game in all its complexity, allowing for larger-scale experimentation to push the boundaries of RL research.

\section{Conclusion and Future Work}

The \nethackenv{} is a fast, complex, procedurally
generated environment for advancing research in RL. We demonstrate that current
state-of-the-art model-free RL serves as a sensible baseline, and
we provide an in-depth analysis of learned agent behaviors.

\nethack{} provides interesting challenges for exploration methods
given the extremely large number of possible states and wide variety
of environment dynamics to discover.
Previously proposed formulations of intrinsic motivation based on
seeking novelty \citep{bellemare2016unifying, ostrovski2017count,
  DBLP:conf/iclr/BurdaESK19} or maximizing surprise
\citep{pathak2017curiosity,
  DBLP:conf/iclr/BurdaEPSDE19,raileanu2020ride} are likely
insufficient to make progress on \nethack{} given that an agent will
constantly find itself in novel states or observe unexpected
environment dynamics.
\nethack{} poses further challenges since, in order to win, an agent
needs to acquire a wide range of skills such as collecting resources,
fighting monsters, eating, manipulating objects, casting spells, or
taking care of their pet, to name just a few.
The multilevel
dependencies present in \nethack{} could inspire progress in
hierarchical RL and long-term planning \citep{Dayan1992FeudalRL,
  Kaelbling1996ReinforcementLA, Parr1997ReinforcementLW,
  Vezhnevets2017FeUdalNF}.  Transfer to unseen game characters,
environment dynamics, or level layouts can be evaluated
\citep{Taylor2009TransferLF}.  Furthermore, its richness and constant
challenge make \nethack{} an interesting benchmark for lifelong
learning \citep{lopez2017gradient, Parisi2019ContinualLL,
  Rolnick2018ExperienceRF, Mankowitz2018UnicornCL}.  In addition, the
extensive documentation about \nethack{} can enable research on using
prior (natural language) knowledge for learning, which
could lead to improvements in generalization and sample efficiency
\citep{Branavan2011LearningTW, luketina2019survey, Zhong2019RTFMGT,
  Jiang2019LanguageAA}.  Lastly, \nethack{} can also drive research on
learning from demonstrations \citep{Abbeel2004ApprenticeshipLV,
  Argall2009ASO} since a large collection of replay data is available.
In sum, we argue that the \nethackenv{} strikes an excellent balance between
complexity and speed while encompassing a variety of challenges for
the research community.

For future versions of the environment, we plan to support NetHack 3.7 once it is released, as it will further increase the variability of observations via \emph{Themed Rooms}. This version will also introduce scripting in the Lua language, which we will leverage to enable users to create their custom sandbox tasks, directly tapping into NetHack and its rich universe of entities and their complex interactions to define custom RL tasks.

% \if 0
\section{Broader Impact}

To bridge the gap between the constrained world of video and board games, and the open and unpredictable real world, there is a need for environments and tasks which challenge the limits of current Reinforcement Learning (RL) approaches.
Some excellent challenges have been put forth over the years, demanding increases in the complexity of policies needed to solve a problem or scale needed to deal with increasingly photorealistic, complex environments.
In contrast, our work seeks to be extremely fast to run while still testing the generalization and exploration abilities of agents in an environment which is rich, procedurally generated, and in which reward is sparse. The impact of solving these problems with minimal environment-specific heuristics lies in the development of RL algorithms which produce sample efficient, robust, and general policies capable of more readily dealing with the uncertain and changing dynamics of ``real world'' environments. We do not solve these problems here, but rather provide the challenge and the testbed against such improvements can be produced and evaluated.

Auxiliary to this, and in line with growing concerns that progress in Deep RL is more the result of industrial labs having privileged access to the resources required to run environments and agents on a massive scale, the environment presented here is computationally cheap to run and to collect data in. This democratizes access for researchers in more resource-constrained labs, while not sacrificing the difficulty and richness of the environment. We hope that as a result of this, and of the more general need to develop sample-efficient agents with fewer data, the environmental impact of research using our environment will be reduced compared to more visually sophisticated ones.
% \fi

% \clearpage
\section*{Acknowledgements}
We thank the NetHack DevTeam for creating and continuously extending
this amazing game over the last decades. We thank Paul Winner, Bart
House, M. Drew Streib, Mikko Juola, Florian Mayer, Philip H.S.~Torr,
Stephen Roller, Minqi Jiang, Vegard Mella, Eric Hambro, Fabio Petroni,
Mikayel Samvelyan, Vitaly Kurin, Arthur Szlam, Sebastian
Riedel, Antoine Bordes, Gabriel Synnaeve, Jeremy Reizenstein, as well
as the NeurIPS 2020, ICML 2020, and BeTR-RL 2020 reviewers and area chairs for their
valuable feedback.
Nantas Nardelli is supported by EPSRC/MURI grant EP/N019474/1.
Finally, we would like to pay tribute to the $863{,}918{,}816$ simulated NetHack heroes who lost their lives in the name of science for this project (thus far).

\bibliography{references}

\clearpage{}

\appendix \onecolumn

\section{Further Details on NetHack}
\label{sec:moreaboutnethack}

\paragraph{Character options}
The player may choose (or pick randomly) the character from thirteen roles (archaeologist, barbarian, cave(wo)man, healer, knight, priest(ess), ranger, rogue, samurai, tourist, valkyrie, and wizard), five races
(human, elf, dwarf, gnome, and orc), three moral alignments (neutral, lawful, chaotic), and two genders
(male or female). Each choice determines
some of the character's features, as well as how the character
interacts with other entities (e.g.,~some species of monsters may not
be hostile depending on the character race; priests of a particular
deity may only help religiously aligned characters).

The hero's interaction with several game entities
involves pre-defined stochastic dynamics (usually defined by virtual
dice tosses), and the game is designed to heavily punish careless
exploration policies.\footnote{Occasionally dying because of simple,
  avoidable mistakes is so common in the game that the online
  community has defined an acronym for it: \emph{Yet Another Stupid
    Death} (YASD).} This makes NetHack an ideal environment
for evaluating exploration methods such as curiosity-driven learning
\citep{pathak2017curiosity, DBLP:conf/iclr/BurdaEPSDE19} or
safe reinforcement learning~\citep{garcia2015comprehensive}.

\begin{sidewaysfigure}[ht]
\centering
    \includegraphics[width=\textwidth]{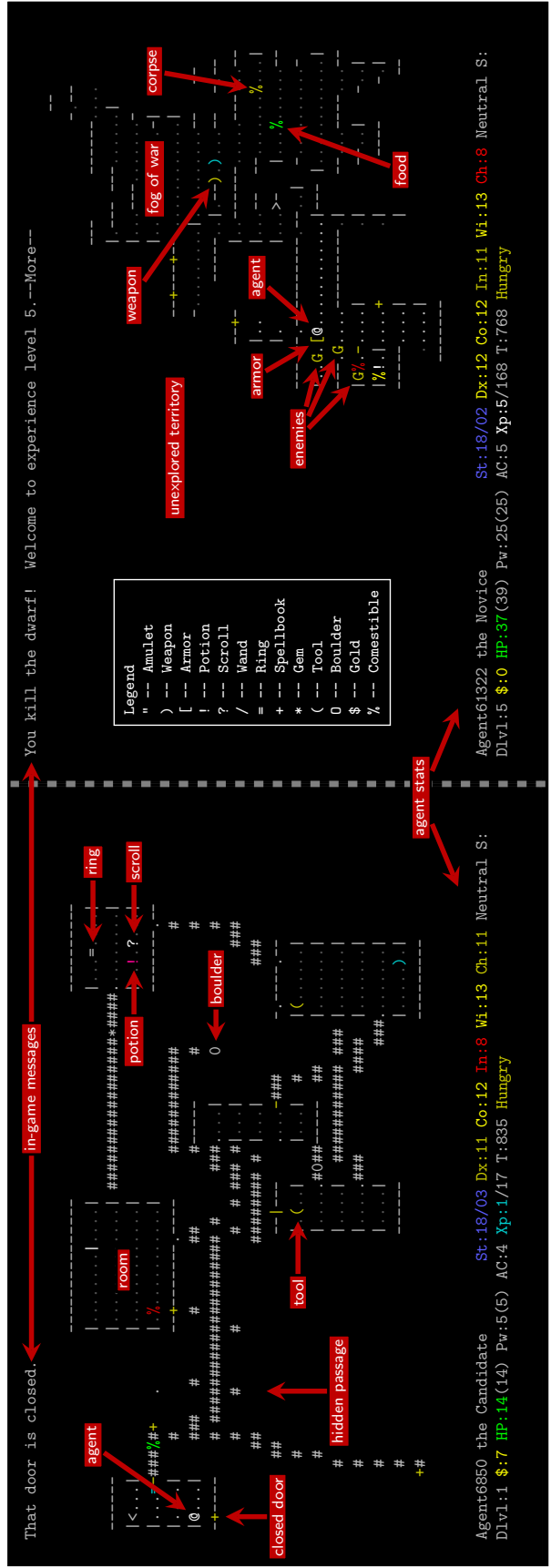}
    \caption{Annotated example of an agent at two different stages in
      \nethack{} (Left: a procedurally generated first level of the
      Dungeons of Doom, right: Gnomish Mines).}
    \label{fig:levelBig}
\end{sidewaysfigure}

Learning and planning in NetHack involves dealing with partial observability. The game, by default, employs \emph{Fog of War} to hide information based on a simple 2D light model (see for example
the difference between white \Tile{} and gray \HiddenTile{} room tiles
in \Cref{fig:level} or \Cref{fig:levelBig}), requiring the player not
only to discover the
topology of the level (including searching for hidden doors and
passages), but to also condition their policy on a world that might
change, e.g., due to monsters spawning and interacting outside of the
visible range.

\begin{center}
\begin{figure}[h]
    \centering
    \includegraphics[width=1\textwidth]{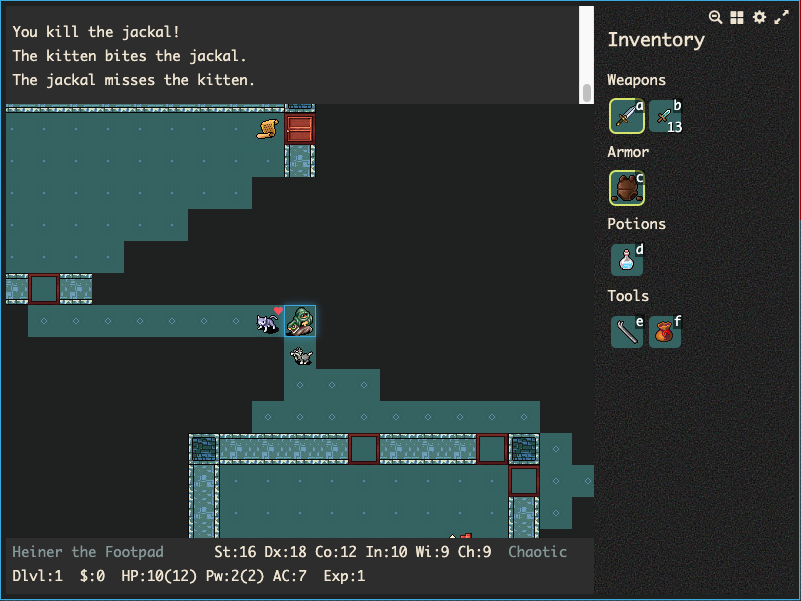}
    \caption{Screenshot of BrowserHack showing NetHack with a graphical user interface.}
    \label{fig:browserhack}
\end{figure}
\end{center}

On top of the standard ASCII interface, \nethack\ supports many official and unofficial graphical user interfaces. Figure~\ref{fig:browserhack} shows a screenshot of Lu Wang's BrowserHack\footnote{Playable online at \url{https://coolwanglu.github.io/BrowserHack/}} as an example.

\paragraph{Conducts}
While winning NetHack by retrieving and ascending with the Amulet of Yendor is already immensely challenging, experienced NetHack players like to challenge themselves even more by imposing additional restrictions on their play.
The game tracks some of these challenges with the \texttt{\#conduct}
command \citep{raymond2020guide}.  These official challenges include eating
only vegan or vegetarian food, or not eating at all, or playing the
game in ``pacifist'' mode without killing a single monster.  While
very experienced players often try to adhere to several challenges at
once, even moderately experienced players often limit their use
of certain polymorph spells (e.g., ``polypiling''---changing the form
of several objects at once in the hope of getting better ones) or they
try to beat the game while \emph{minimizing} the in-game score.  We
believe this established set of conducts will supply the RL community
with a steady stream of extended challenges once the standard \nethackenv{} is solved by future methods.

\section{Observation Space}
\label{sec:observation}

The Gym environment is implemented by wrapping a more
low-level \texttt{NetHack} Python object into a Python class
responsible for the featurization, reward schedule and end-of-episode
dynamics. While the low-level \texttt{NetHack} object gives access to a large
number of NetHack game internals, the Gym wrapper exposes by default
only a part of this data as numerical observation arrays, namely the
observation tensors
\textit{glyphs}, \textit{chars}, \textit{colors}, \textit{specials},
\textit{blstats}, \textit{message}, \textit{inv\_glyphs}, \textit{inv\_strs},
\textit{inv\_letters}, and \textit{inv\_oclasses}.

\paragraph{Glyphs, Chars, Colors, Specials:}  NetHack supports
non-ASCII graphical user interfaces, dubbed window-ports
(see \Cref{fig:browserhack} for an example). To support displaying
different monsters, objects and floor types in the NetHack dungeon map
as different tiles, NetHack internally defines \emph{glyphs} as ids in
the range $0, \ldots, \texttt{MAX\_GLYPH}$, where $\texttt{MAX\_GLYPH}
= 5991$ in our build\footnote{The exact number of monsters in NetHack
depends on compile-time options as well as the target operating
system. For instance, the mail daemon \protect\MailDaemon\ is only available
on Unix-like operating systems, where it delivers
email in the form of a NetHack scroll if the system is configured to
host a Unix mailbox.}. The \textit{glyph} observation is an integer
array of shape $(21, 79)$ of these game glyph ids.\footnote{NetHack's
  set of glyph ids is not necessarily well suited for machine
  learning. For example, more than half of all glyph ids are of type
  ``swallow'', most of which are guaranteed not to show up in any
  actual game of NetHack. We provide additional tooling to determine
  the type of a given glyph id to process this observation further.}
In NetHack's standard terminal-based user interface, these glyphs are
mapped into ASCII characters of different colors which we return as
the \textit{chars}, \textit{colors}, and \textit{specials}
observations, both all which are
of shape $(21, 79)$; \textit{chars} are ASCII bytes in the range $0,\ldots,127$
wheras \textit{colors} are in range $0,\ldots,15$. For additional
highlighting (e.g., flipping background and foreground colors for the
hero's pet),
NetHack also computes xor'ed values which we return as
the \textit{specials} tensor.

\paragraph{Blstats:} ``Bottom line statistics'', a integer vector of
length~25, containing
the $(x, y)$ coordinate of the hero and the following $23$ character
stats that typically appear in the bottom line of the ASCII
interface: \texttt{strength\_percentage}, \texttt{strength},
\texttt{dexterity}, \texttt{constitution}, \texttt{intelligence},
\texttt{wisdom}, \texttt{charisma}, \texttt{score},
\texttt{hitpoints}, \texttt{max\_hitpoints},
\texttt{depth},
\texttt{gold},
\texttt{energy}, \texttt{max\_energy}, \texttt{armor\_class},
\texttt{monster\_level}, \texttt{experience\_level},
\texttt{experience\_points}, \texttt{time}, \texttt{hunger\_state},
\texttt{carrying\_capacity}, \texttt{dungeon\_number}, and \texttt{level\_number}.

\paragraph{Message:} A padded byte vector of length $256$
representing the current message shown to the player, normally
displayed in the top area of the GUI. We support different padding
strategies and alphabet sizes, but by default we choose an alphabet
size of $96$, where the last character is used for padding.

\paragraph{Inventory:} In NetHack's default ASCII user
interface, the hero's inventory can be opened and closed during the
game. Other user interfaces display a permanent inventory at all
times. \NLE{} follows that strategy. The inventory observations
consist of the following four arrays: \textit{inv\_glyphs}: an integer
vector of length $55$ of glyph ids, padded with
$\texttt{MAX\_GLYPH}$; \ \textit{inv\_strs}: A padded byte array of
shape $(55, 80)$ describing the inventory
items; \ \textit{inv\_letters}: A padded byte vector of length $55$
with the corresponding ASCII character
symbol; \ \textit{inv\_oclasses}: An integer vector of shape $55$ with
ids describing the type of inventory objects, padded with
$\texttt{MAXOCLASSES} = 18$.

The low-level \texttt{NetHack} Python object has some additional
methods to query and modify NetHack's game state, e.g. the current RNG
seeds. We refer to the source code to describe these.\footnote{See,
e.g., the \texttt{nethack.py} as well as \texttt{pynethack.cc} files
in the \NLE\ repository.}

\section{Action Space}
\label{sec:actions}

The game of \nethack\ uses ASCII inputs, i.e., individual keyboard
presses including modifiers like Ctrl and Meta. \NLE{}
pre-defines~$98$ actions, $16$ of which are compass
directions and $82$ of which are command
actions. Table~\ref{table:commandactions} gives a list of command
actions, including their ASCII value and the corresponding key binding
in \nethack{}, while Table~\ref{table:compassdirections} lists the
16~compass directions. For a detailed description of these actions, as
well as other \nethack\ commands, we refer the reader to the NetHack guide book
\citep{raymond2020guide}. Not all actions are sensible for standard RL
training on \NLE. E.g., the \textsc{version} or \textsc{quit} actions
are unlikely to be useful for direct input from the agent. \NLE\
defines a list of \texttt{USEFUL\_ACTIONS} that includes a subset of
76~actions; however, what is useful depends on the circumstances. In
addition, even though an action like \textsc{save} is unlikely to be
useful in most game situations it corresponds to the
letter \texttt{S}, which may be assigned to an inventory item or some
other in-game menu entry such that it does become a useful action in
that context.

By default, \NLE\ will auto-apply the \textsc{more} action in
situations where the game waits for input to display more
messages.
% Hack to get footnote in the table below.
\addtocounter{footnote}{1}
\footnotetext{The descriptions are mostly taken from the
  \texttt{cmd.c} file in the \nethack\ source code.}

\begin{center}
  \captionsetup{type=table}
  \captionof{table}{Command actions.\protect\footnotemark[\thefootnote]}%
  \small
\begin{longtable}{>{\scshape}lr>{\ttfamily}ll}
  % See also https://docs.google.com/spreadsheets/d/1WUyEsi3Yv_sZi0EN66_ltx9QsekypWipmhlftGhiJqk/edit?usp=sharing
  \toprule \\
    \textnormal{Name} & \textnormal{Value} & \textnormal{Key} &
    Description \tabularnewline
    \midrule
    extcmd & 35 & \#  & perform an extended command\\
    extlist & 191 & M-?  & list all extended commands\\
    adjust & 225 & M-a  & adjust inventory letters\\
    annotate & 193 & M-A  & name current level\\
    apply & 97 & a  & apply (use) a tool (pick-axe, key, lamp...)\\
    attributes & 24 & C-x  & show your attributes\\
    autopickup & 64 & @  & toggle the pickup option on/off\\
    call & 67 & C  & call (name) something\\
    cast & 90 & Z  & zap (cast) a spell\\
    chat & 227 & M-c  & talk to someone\\
    close & 99 & c  & close a door\\
    conduct & 195 & M-C  & list voluntary challenges you have maintained\\
    dip & 228 & M-d  & dip an object into something\\
    down & 62 & \textgreater & go down (e.g., a staircase) \\
    drop & 100 & d  & drop an item\\
    droptype & 68 & D  & drop specific item types\\
    eat & 101 & e  & eat something\\
    esc & 27 & C-[ & escape from the current query/action\\
    engrave & 69 & E  & engrave writing on the floor\\
    enhance & 229 & M-e  & advance or check weapon and spell skills\\
    fire & 102 & f  & fire ammunition from quiver\\
    fight & 70 & F  & Prefix: force fight even if you don't see a monster\\
    force & 230 & M-f  & force a lock\\
    glance & 59 & ;  & show what type of thing a map symbol corresponds to\\
    help & 63 & ?  & give a help message\\
    history & 86 & V  & show long version and game history\\
    inventory & 105 & i  & show your inventory\\
    inventtype & 73 & I  & inventory specific item types\\
    invoke & 233 & M-i  & invoke an object's special powers\\
    jump & 234 & M-j  & jump to another location\\
    kick & 4 & C-d  & kick something\\
    known & 92 & \textbackslash  & show what object types have been discovered\\
    knownclass & 96 & `  & show discovered types for one class of objects\\
    look & 58 & :  & look at what is here\\
    loot & 236 & M-l  & loot a box on the floor\\
    monster & 237 & M-m  & use monster's special ability\\
    more & 13 & C-m  & read the next message\\
    move & 109 & m  & Prefix: move without picking up objects/fighting\\
    movefar & 77 & M & Prefix: run without picking up objects/fighting\\
    % name & 78 & N  & name a monster or an object\\  % clashes with SE-long below.
    offer & 239 & M-o  & offer a sacrifice to the gods\\
    open & 111 & o  & open a door\\
    options & 79 & O  & show option settings, possibly change them\\
    overview & 15 & C-o  & show a summary of the explored dungeon\\
    pay & 112 & p  & pay your shopping bill\\
    pickup & 44 & ,  & pick up things at the current location\\
    pray & 240 & M-p  & pray to the gods for help\\
    prevmsg & 16 & C-p  & view recent game messages\\
    puton & 80 & P  & put on an accessory (ring, amulet, etc)\\
    quaff & 113 & q  & quaff (drink) something\\
    quit & 241 & M-q  & exit without saving current game\\
    quiver & 81 & Q  & select ammunition for quiver\\
    read & 114 & r  & read a scroll or spellbook\\
    redraw & 18 & C-r  & redraw screen\\
    remove & 82 & R  & remove an accessory (ring, amulet, etc)\\
    ride & 210 & M-R  & mount or dismount a saddled steed\\
    rub & 242 & M-r  & rub a lamp or a stone\\
    rush & 103 & g  & Prefix: rush until something interesting is seen\\
    save & 83 & S  & save the game and exit\\
    search & 115 & s  & search for traps and secret doors\\
    seeall & 42 & *  & show all equipment in use\\
    seetrap & 94 & \textasciicircum  & show the type of adjacent trap\\
    sit & 243 & M-s  & sit down\\
    swap & 120 & x  & swap wielded and secondary weapons\\
    takeoff & 84 & T  & take off one piece of armor\\
    takeoffall & 65 & A  & remove all armor\\
    teleport & 20 & C-t  & teleport around the level\\
    throw & 116 & t  & throw something\\
    tip & 212 & M-T  & empty a container\\
    travel & 95 & \_  & travel to a specific location on the map\\
    turn & 244 & M-t  & turn undead away\\
    twoweapon & 88 & X  & toggle two-weapon combat\\
    untrap & 245 & M-u  & untrap something\\
    up & 60 & \textless & go up (e.g., a staircase) \\
    version & 246 & M-v  & list compile time options\\
    versionshort & 118 & v  & show version\\
    wait / self & 46 & . & rest one move while doing nothing / apply to self \\
    wear & 87 & W  & wear a piece of armor\\
    whatdoes & 38 & \&  & tell what a command does\\
    whatis & 47 & /  & show what type of thing a symbol corresponds to\\
    wield & 119 & w  & wield (put in use) a weapon\\
    wipe & 247 & M-w  & wipe off your face\\
    zap & 112 & z  & zap a wand\\
  \bottomrule%
\end{longtable}%
\normalsize%

  \label{table:commandactions}
\end{center}

\begin{table}[h]
\centering
\caption{Compass direction actions.\\[1em]}
\label{table:compassdirections}
\begin{tabular}{lc>{\ttfamily}cc>{\ttfamily}c}
\toprule & \multicolumn{2}{c}{one-step} & \multicolumn{2}{c}{move far}
\\ \cmidrule(l){2-3} \cmidrule(l){4-5} Direction & \textnormal{Value}
& \textnormal{Key} & Value & \textnormal{Key} \tabularnewline \midrule
North & 107 & k & 75 & K \\ East & 108 & l & 76 & L \\ South & 106 & j
& 74 & J \\ West & 104 & h & 72 & H \\ North East & 117 & u & 85 & U
\\ South East & 110 & n & 78 & N \\ South West & 98 & b & 66 & B
\\ North West & 121 & y & 89 & Y \\ \bottomrule
\end{tabular}
\end{table}

\section{Environment Speed Comparison}
\label{sec:envsspeed}

Table~\ref{table:envscomparison} shows a comparison between popular Gym
environments and \NLE{}.
All environments were controlled with a
uniformly random policy using reset on terminal states.
The tests were conducted on a MacBook
Pro equipped with an Intel Core i7 2.9 GHz, 16GB of RAM, MacOS Mojave,
Python 3.7, Conda 4.7.12, and latest available packages as of May
2020. \emph{ObstacleTowerEnv} was instantiated with \texttt{(retro=False, real\_time=False)}. Note that this data does not necessarily reflect performance of these environments with better---or worse---policies, as each environment has computational dynamics that depend on its state. However, we expect the difference in terms of magnitude to remain mostly unchanged across these environments.
\begin{table}[h]
\centering
\caption{Comparison between \NLE\ and popular environments when
   using their respective Python Gym interface. SPS stands for ``environment steps per
  second''. All environments but \texttt{ObstacleTowerEnv} were run via \texttt{gym} with standard settings (and headless when possible), for $60$ seconds.\\}
\label{table:envscomparison}
\begin{tabular}{lrrr}
\toprule
Environment &    SPS &     steps & episodes \\
\midrule
                        \NLE\ (score)   &  14.4K &   868.75K &      477 \\
                          CartPole-v1   & 76.88K &  4612.65K &   207390 \\
 ALE (MontezumaRevengeNoFrameskip-v4)   &  0.90K &    53.91K &      611 \\
          Retro (Airstriker-Genesis)    &  1.31K &    78.56K &       52 \\
           ProcGen (procgen-coinrun-v0) & 13.13K &   787.98K &     1283 \\
                     ObstacleTowerEnv   &  0.06K &     3.61K &        6 \\
               MineRLNavigateDense-v0   &  0.06K &     3.39K &        0 \\
\bottomrule
\end{tabular}

\end{table}

\section{Task Details}
\label{sec:taskdetails}
\begin{figure}
  \centering
  \includegraphics[width=0.6\linewidth]{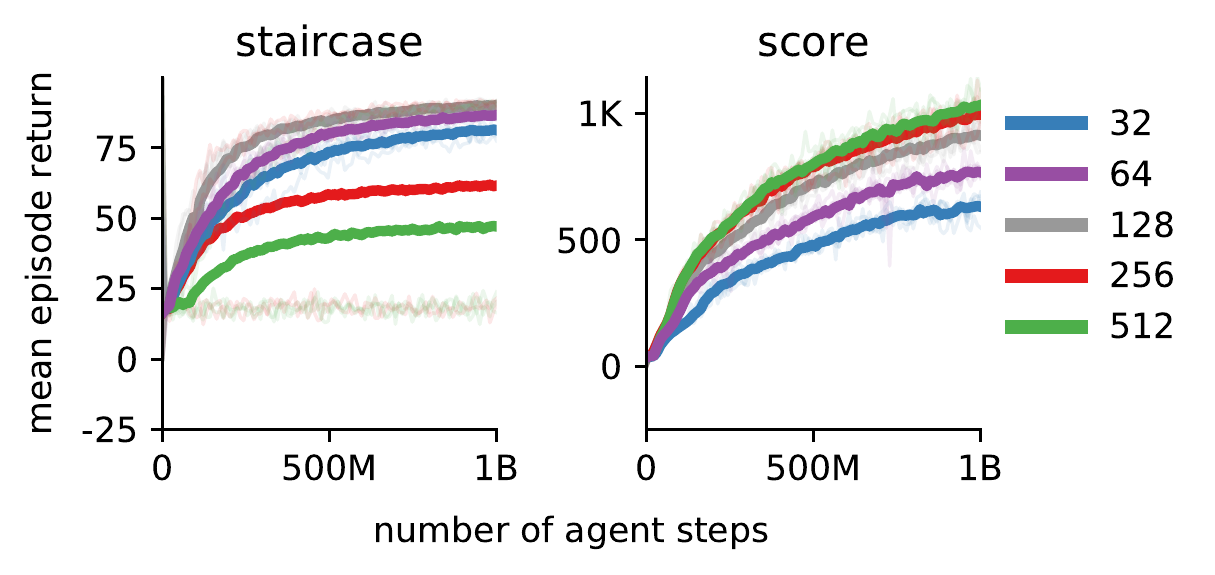}
  \captionof{figure}{Mean episode return of the last $100$ episodes for models with different hidden sizes averaged over five runs.}
  \label{fig:results_size}
\end{figure}

For all tasks described below, we add a penalty of $-0.001$ to the reward function if the agent's action did not advance the in-game timer, which, for example, happens when the agent tries to move against a wall or
navigates menus.  For all tasks, except the \emph{Gold} task, we
disable \nethack{}'s \emph{autopick} option \citep{raymond2020guide}.
Furthermore, we disable so-called \emph{bones files} that would
otherwise lead to agents occasionally discovering the remains and
ghosts of previous agents, considerably increasing the variance across
episodes.

\paragraph{Staircase}
The agent has to find the staircase down \StaircaseDown{} to the
next dungeon level.  This task is already challenging, as there is
often no direct path to the staircase.  Instead, the agent has to
learn to reliably open doors \Door{}, kick-in locked doors, search for
hidden doors and passages \Passage{}, avoid traps \Trap{}, or move
boulders \Boulder{} that obstruct a passage.  The agent receives a
reward of $100$ once it reaches the staircase down and the the episode
terminates after $1000$ agent steps.
\paragraph{Pet}
Many successful strategies for \nethack{} rely on taking good care of
the hero's pet (e.g., the little dog \LittleDog{} or kitten
\Kitten{} that the hero starts with).  Pets are controlled by the game,
but their behavior is influenced by the agent's actions.  In this
task, the agent only receives a positive reward of 100 when it reaches the
staircase while the pet is next to the agent.
\paragraph{Eat}
To survive in \nethack{}, players have to make sure their character
does not starve to death.  There are many edible objects in the game,
for example food rations \Food{}, tins, and monster corpses.  In this
task, the agent receives the increase of nutrition as determined by
the in-game ``Hunger'' status as reward \citep[see][``Nutrition''
  entry for details]{nhwiki}.  A steady source of nutrition are
monster corpses, but for that the agent has to learn to locate and to kill
monsters while avoiding to consume rotten corpses, poisonous monster corpses
such as Kobolds \Kobold{} or acidic monster corpses such as Acid Blobs
\AcidBlob{}.

\paragraph{Gold}
Throughout the game, the player can collect gold \Gold{} to, for example, trade for
useful items with shopkeepers.  The agent receives the amount of gold
it collects as reward.  This incentivizes the agent to explore dungeon
maps fully and to descend dungeon levels to discover new sources of
gold.  There are many advanced strategies for obtaining large amounts
of gold such as finding, identifying and selling gems; stealing from
or killing shopkeepers; or hunting for vaults or leprechaun halls.
To make this task easier for the agent, we enable \nethack{}'s
\emph{autopickup} option for gold.

\paragraph{Scout}
An important part of the game is exploring dungeon levels. Here, we
reward the agent ($+1$) for uncovering previously unknown tiles in the
dungeon, for example by entering a new room or following a newly
discovered passage.  Like the previous task, this incentivizes the
agent to explore dungeon levels and to descend.

\paragraph{Score}
In this task, the agent receives the increase of the in-game
score between two time steps as reward.
The in-game score is governed by a complex calculation, but in
early stages of the game it is dominated by killing monsters and the number of
dungeon levels that the agent descends \citep[see][``Score'' entry for
  details]{nhwiki}.

\paragraph{Oracle}
While levels are procedurally generated, there are a number of
landmarks that appear in every game of \nethack{}.  One such landmark
is the Oracle \Oracle{}, which is randomly placed between levels five and nine
of the dungeon.  Reliably finding the Oracle is difficult, as it
requires the agent to go down multiple staircases and often to
exhaustively explore each level.  In this task, the agent receives a
reward of $1000$ if it manages to reach the Oracle.

\section{Baseline CNN Details}
\label{appendix:baseline}
As embedding dimension of the glyphs we use $32$ and for the hidden dimension for
the observation $\mathbf{o}_t$ and the output of the LSTM $\mathbf{h}_t$,
we use $128$.
For encoding the full map of glyphs as well as the $9\times 9$ crop,
we use a $5$-layer ConvNet architecture with filter size $3\times 3$,
padding $1$ and stride $1$. The input channel of the first layer of
the ConvNet is the embedding size of the glyphs ($32$). Subsequent
layers have an input and output channel dimension of $16$.
We employ a gradient norm clipping of $40$ and clip rewards using $r_c
= \tanh(r / 100)$.
We use RMSProp with a learning rate of $0.0002$ without momentum and
with $\varepsilon_{\textrm{RMSProp}} = 0.000001$. Our entropy cost is set
to $0.0001$.

\section{Random Network Distillation Details}
\label{appendix:rnd}
For RND hyperparameters we mostly follow the recommendations by the authors~\citep{DBLP:conf/iclr/BurdaESK19}:
\begin{itemize}
    \item we initialize the weights according to the original paper, using an orthogonal distribution with a gain of $\sqrt{2}$
    \item we use a two-headed value function rather than merely summing the intrinsic and extrinsic reward
    \item we use a discounting factor of $0.999$ for the extrinsic reward and $0.99$ for the intrinsic reward
    \item we use non-episodic intrinsic reward and episodic extrinsic reward
    \item we use reward normalization for the intrinsic reward, dividing it by a running estimate of its standard deviation
\end{itemize}

We modify a few of the parameters for use in our setting:
\begin{itemize}
    \item we use exactly the same feature extraction architecture as the baseline model instead of the pixel-based convolutional feature extractor
    \item we do not use observation normalization, again due to the symbolic nature of our observation space
    \item before normalizing, we divide the intrinsic reward by ten so that it has less weight than the extrinsic reward
    \item we clip intrinsic rewards in the same way that we clip extrinsic rewards, i.e., using $r_c = \tanh(r / 100)$, so that the intrinsic and extrinsic rewards are on a similar scale
\end{itemize}

We downscale the forward modeling loss by a factor of $0.01$ to slow down the rate at which the model becomes familiar with a given state, since the intrinsic reward often collapsed quickly despite the reward normalization.
We determined these settings during a set of small-scale experiments.

We also tried using subsets of the full feature set (only the embedding of the full display of glyphs, or only the embedding of the crop of glyphs around the agent) as well as the exact architecture used by the original authors, but with the pixel input replaced by a random $8$-dimensional embedding of the symbolic observation space. However, we did not observe this improved results.

We tried using intrinsic reward only as the authors did in the original RND paper, but we found that agents trained in this way made no significant progress through the dungeon, even on a single fixed seed. This indicates that this form of intrinsic reward is not sufficient to make progress on NetHack.
As noted in \Cref{sec:experiments}, the intrinsic reward did help in some tasks for some characters when combined with the extrinsic reward. Crucially, RND exploration is not sufficient for agents to learn to find the Oracle, which leaves this as a difficult challenge for future exploration techniques.

\section{Dashboard}
\label{sec:dashboard}
\begin{sidewaysfigure}[ht]
\centering
    \includegraphics[width=\textwidth]{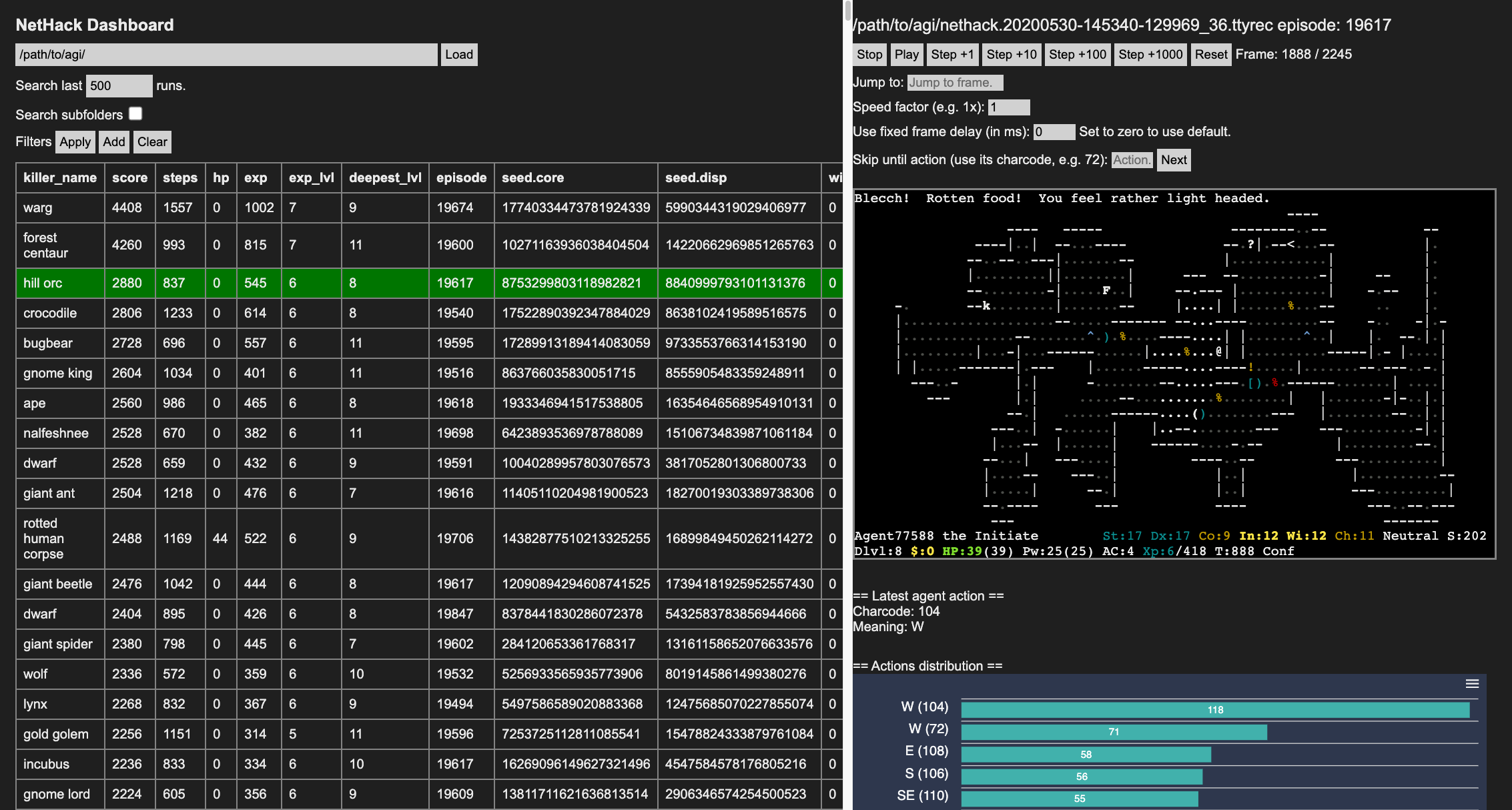}
    \caption{Screenshot of the web dashboard included in the \nethackenv{}.}
    \label{fig:dashboard}
\end{sidewaysfigure}

We release a web dashboard built with NodeJS (see \Cref{fig:dashboard}) to visualize experiment runs and statistics for \NLE{}, including replaying episodes that were recorded as \texttt{tty} files.

\section{NetHack Bots}
\label{sec:bots}
% add https://en.wikipedia.org/wiki/Rog-O-Matic ?
Since the early stages of the development of NetHack, players have tried to build bots to play and solve the game.
Notable examples are \emph{TAEB}, \emph{BotHack}, and
\emph{Saiph}~\citep{taebwikibots, nhwiki}. These bot frameworks largely rely on search heuristics and common planning methods, without generally making use of any statistical learning methods. An exception is \emph{SWAGGINZZZ}~\citep{swagginzzz} which uses lookups, exhaustive simulation and manipulation of the random number generator.

Successful bots have made use of exploits that are no longer present in recent versions of NetHack. For example, \emph{BotHack} employs the ``pudding farming'' strategy \citep[see][``Pudding farming'' entry]{nhwiki} to level up and to create items for the character by spawning and killing a large number of black puddings~\BlackPudding{}. This enabled the bot to become quite strong, which rendered late-game fights considerably easier. This strategy was disabled by the \nethack{} DevTeam with a patch that is incorporated into versions of \nethack{} above 3.6.0. Likewise, the random number generator manipulations employed in \emph{SWAGGINZZZ} are no longer possible.

We believe that it is very unlikely that in the future we will see a hand-crafted bot solving NetHack in the way we defined it in \Cref{sec:eval}.
In fact, the creator of \emph{SWAGGINZZZ} remarked that ``[e]ven with RNG manipulation, writing a bot that 99\% ascends NetHack is \textbf{extremely} complicated. So much stuff can go wrong, and there is no shortage of corner cases'' \citep{swagginzzz}.

\section{Viewing Agent Videos}
\label{sec:runsamples}

We have uploaded some agent recordings to \url{https://asciinema.org/~nle}. These can be either watched on the Asciinema portal, or on a terminal by running \texttt{asciinema play -s 0.2 \textit{url}} (\texttt{asciinema} itself is available as a pip package at \url{https://pypi.org/project/asciinema}). The \texttt{-s} flag regulates the speed of the recordings, which can also be modified on the web interface by pressing \texttt{>} (faster) or \texttt{<} (slower).

\begin{figure*}[t!]
    \centering \includegraphics[width=1.0\textwidth]{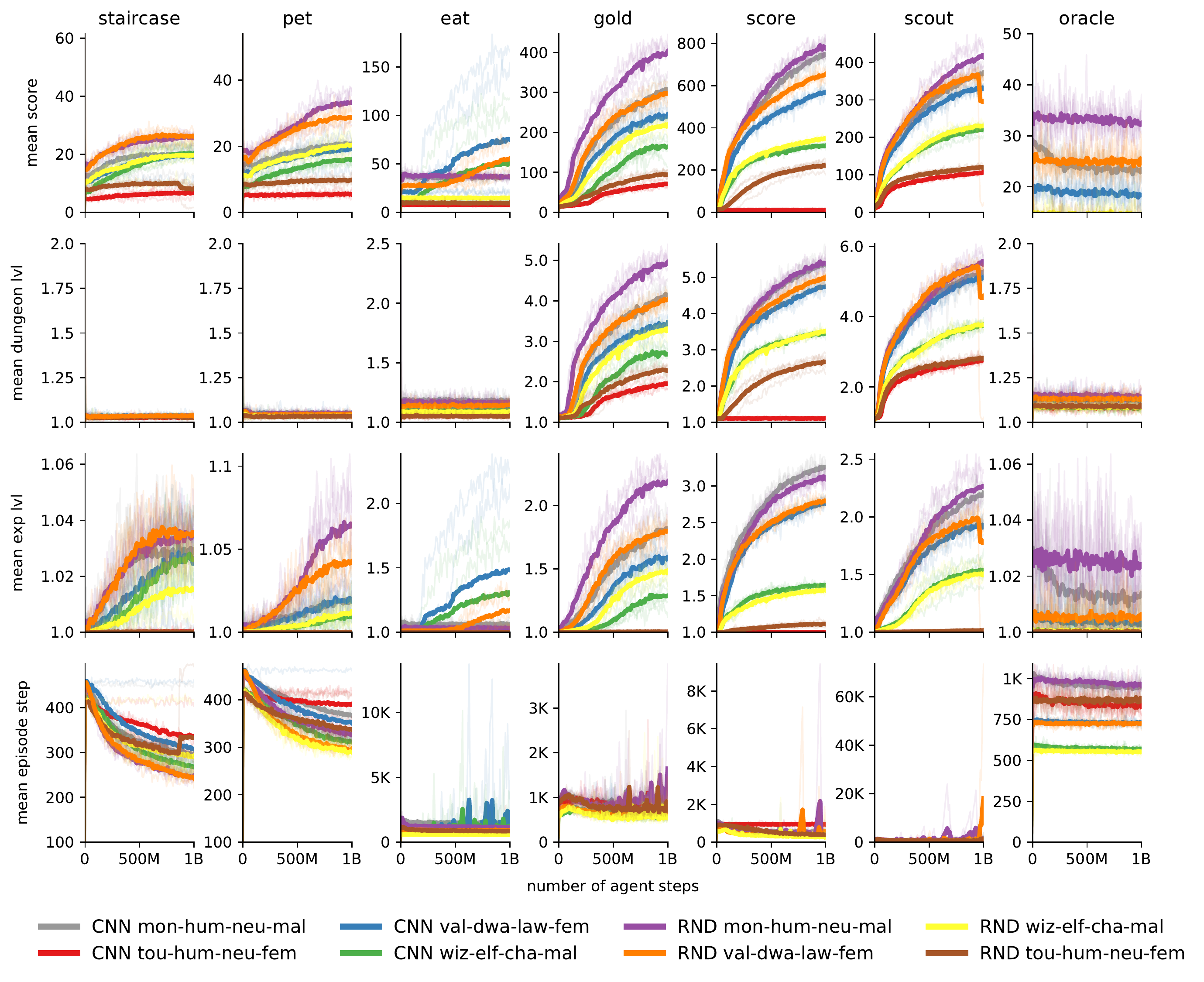}
    \caption{Mean score, dungeon level reached, experience level achieved, and steps performed in the environment in the last $100$ episodes averaged over five runs.}
    \label{fig:eval_extra}
\end{figure*}

\begin{table}[t!]
    \centering
\caption{Metrics averaged over last $1000$ episodes for each task.\\[1em]}
\resizebox{0.85\linewidth}{!}{
\begin{tabular}{lllrrrrr}
\toprule
      Task & Model &        Character &  Score &  Time &  Exp Lvl &  Dungeon Lvl &    Win \\
\midrule
 staircase &   CNN &  mon-hum-neu-mal &     20 &   252 &               1.0 &            1.0 &  77.26 \\
  &    &  tou-hum-neu-fem &      6 &   288 &               1.0 &            1.0 &  50.42 \\
  &    &  val-dwa-law-fem &     19 &   329 &               1.0 &            1.0 &  74.62 \\
  &    &  wiz-elf-cha-mal &     20 &   253 &               1.0 &            1.0 &  80.42 \\
\cmidrule{2-8}
  &   RND &  mon-hum-neu-mal &     26 &   199 &               1.0 &            1.0 &  90.84 \\
  &    &  tou-hum-neu-fem &      8 &   203 &               1.0 &            1.0 &  56.94 \\
  &    &  val-dwa-law-fem &     25 &   242 &               1.0 &            1.0 &  90.96 \\
  &    &  wiz-elf-cha-mal &     20 &   317 &               1.0 &            1.0 &  67.46 \\
\midrule
       pet &   CNN &  mon-hum-neu-mal &     20 &   297 &               1.0 &            1.1 &  62.02 \\
        &    &  tou-hum-neu-fem &      6 &   407 &               1.0 &            1.0 &  25.66 \\
        &    &  val-dwa-law-fem &     18 &   379 &               1.0 &            1.0 &  63.30 \\
        &    &  wiz-elf-cha-mal &     16 &   273 &               1.0 &            1.0 &  66.80 \\
\cmidrule{2-8}
        &   RND &  mon-hum-neu-mal &     33 &   319 &               1.1 &            1.0 &  74.38 \\
        &    &  tou-hum-neu-fem &     10 &   336 &               1.0 &            1.0 &  49.38 \\
        &    &  val-dwa-law-fem &     28 &   311 &               1.0 &            1.0 &  81.56 \\
        &    &  wiz-elf-cha-mal &     20 &   278 &               1.0 &            1.0 &  70.48 \\
\midrule
       eat &   CNN &  mon-hum-neu-mal &     36 &  1254 &               1.1 &            1.2 &   -- \\
        &    &  tou-hum-neu-fem &      7 &   423 &               1.0 &            1.0 &   -- \\
        &    &  val-dwa-law-fem &     75 &  1713 &               1.5 &            1.1 &   -- \\
        &    &  wiz-elf-cha-mal &     50 &  1181 &               1.3 &            1.1 &   -- \\
\cmidrule{2-8}
        &   RND &  mon-hum-neu-mal &     36 &  1102 &               1.0 &            1.2 &   -- \\
        &    &  tou-hum-neu-fem &      9 &   404 &               1.0 &            1.0 &   -- \\
        &    &  val-dwa-law-fem &     55 &  1421 &               1.2 &            1.1 &   -- \\
        &    &  wiz-elf-cha-mal &     14 &   808 &               1.0 &            1.1 &   -- \\
\midrule
      gold &   CNN &  mon-hum-neu-mal &    307 &   947 &               1.8 &            4.2 &   -- \\
       &    &  tou-hum-neu-fem &     71 &   788 &               1.0 &            2.0 &   -- \\
       &    &  val-dwa-law-fem &    245 &  1032 &               1.6 &            3.5 &   -- \\
       &    &  wiz-elf-cha-mal &    162 &   780 &               1.3 &            2.7 &   -- \\
\cmidrule{2-8}
       &   RND &  mon-hum-neu-mal &    403 &  1006 &               2.2 &            5.0 &   -- \\
       &    &  tou-hum-neu-fem &     92 &   816 &               1.0 &            2.2 &   -- \\
       &    &  val-dwa-law-fem &    298 &   998 &               1.8 &            4.0 &   -- \\
       &    &  wiz-elf-cha-mal &    217 &   789 &               1.5 &            3.3 &   -- \\
\midrule
     score &   CNN &  mon-hum-neu-mal &    748 &   932 &               3.2 &            5.4 &   -- \\
      &    &  tou-hum-neu-fem &     11 &   795 &               1.0 &            1.1 &   -- \\
      &    &  val-dwa-law-fem &    573 &   908 &               2.8 &            4.8 &   -- \\
      &    &  wiz-elf-cha-mal &    314 &   615 &               1.6 &            3.5 &   -- \\
\cmidrule{2-8}
      &   RND &  mon-hum-neu-mal &    780 &   863 &               3.1 &            5.4 &   -- \\
      &    &  tou-hum-neu-fem &    219 &   490 &               1.1 &            2.6 &   -- \\
      &    &  val-dwa-law-fem &    647 &   857 &               2.8 &            5.0 &   -- \\
      &    &  wiz-elf-cha-mal &    352 &   585 &               1.6 &            3.5 &   -- \\
\midrule
     scout &   CNN &  mon-hum-neu-mal &    372 &   838 &               2.2 &            5.3 &   -- \\
      &    &  tou-hum-neu-fem &    105 &   580 &               1.0 &            2.7 &   -- \\
      &    &  val-dwa-law-fem &    331 &   852 &               1.9 &            5.1 &   -- \\
      &    &  wiz-elf-cha-mal &    222 &   735 &               1.5 &            3.8 &   -- \\
\cmidrule{2-8}
      &   RND &  mon-hum-neu-mal &    416 &   924 &               2.3 &            5.5 &   -- \\
      &    &  tou-hum-neu-fem &    119 &   599 &               1.0 &            2.8 &   -- \\
      &    &  val-dwa-law-fem &    304 &  1021 &               1.8 &            4.6 &   -- \\
      &    &  wiz-elf-cha-mal &    231 &   719 &               1.5 &            3.8 &   -- \\
\midrule
    oracle &   CNN &  mon-hum-neu-mal &     24 &   876 &               1.0 &            1.1 &   0.00 \\
     &    &  tou-hum-neu-fem &      9 &   674 &               1.0 &            1.1 &   0.00 \\
     &    &  val-dwa-law-fem &     18 &  1323 &               1.0 &            1.1 &   0.02 \\
     &    &  wiz-elf-cha-mal &     10 &   742 &               1.0 &            1.1 &   0.00 \\
\cmidrule{2-8}
     &   RND &  mon-hum-neu-mal &     32 &   967 &               1.0 &            1.1 &   0.00 \\
     &    &  tou-hum-neu-fem &     13 &   811 &               1.0 &            1.1 &   0.00 \\
     &    &  val-dwa-law-fem &     26 &  1353 &               1.0 &            1.1 &   0.00 \\
     &    &  wiz-elf-cha-mal &     14 &   791 &               1.0 &            1.1 &   0.00 \\
\bottomrule
\end{tabular}
}
\label{tab:results:score}
\end{table}

\begin{table}
  \centering
  \caption{Top five of the last $1000$ episodes in the score task.\\[1em]}
\resizebox{0.95\linewidth}{!}{
\begin{tabular}{lllrrr}
\toprule
Model &        Character &                           Killer Name &  Score &  Exp Lvl &  Dungeon Lvl \\
\midrule
  CNN &  mon-hum-neu-mal &                                  warg &   4408 &                 7 &              9 \\
   &   &                        forest centaur &   4260 &                 7 &             11 \\
   &   &                              hill orc &   2880 &                 6 &              8 \\
   &   &                            gnome lord &   2848 &                 6 &              9 \\
   &   &                             crocodile &   2806 &                 6 &              8 \\
   \cmidrule{2-6}
   &  tou-hum-neu-fem &                                jackal &    200 &                 1 &              3 \\
   &   &                             hobgoblin &    200 &                 1 &              5 \\
   &   &                                hobbit &    200 &                 1 &              3 \\
   &   &                             giant rat &    190 &                 1 &              4 \\
   &   &                          large kobold &    174 &                 1 &              4 \\
   \cmidrule{2-6}
   &  val-dwa-law-fem &                            gnome lord &   2176 &                 5 &             12 \\
   &   &                                   ape &   1948 &                 6 &              7 \\
   &   &                               gremlin &   1924 &                 5 &             11 \\
   &   &                            gnome king &   1916 &                 5 &             11 \\
   &   &                               vampire &   1864 &                 4 &             10 \\
   \cmidrule{2-6}
   &  wiz-elf-cha-mal &                                 dingo &   1104 &                 3 &              9 \\
   &   &                             giant ant &   1008 &                 3 &              8 \\
   &   &                           gnome mummy &    988 &                 3 &              8 \\
   &   &                                coyote &    988 &                 3 &              9 \\
   &   &                        kicking a wall &    972 &                 3 &              8 \\
\midrule
  RND &  mon-hum-neu-mal &                                 rothe &   3664 &                 5 &              7 \\
   &   &                   rotted dwarf corpse &   3206 &                 5 &              7 \\
   &   &                             leocrotta &   2771 &                 5 &             11 \\
   &   &                       winter wolf cub &   2724 &                 6 &              9 \\
   &   &                            starvation &   2718 &                 6 &              6 \\
   \cmidrule{2-6}
   &  tou-hum-neu-fem &                              grid bug &   1432 &                 1 &              7 \\
   &   &                             sewer rat &   1253 &                 1 &              4 \\
   &   &                          bolt of cold &   1248 &                 1 &              3 \\
   &   &                                goblin &   1125 &                 1 &              4 \\
   &   &                                goblin &   1078 &                 1 &              4 \\
   \cmidrule{2-6}
   &  val-dwa-law-fem &                               bugbear &   2186 &                 6 &              9 \\
   &   &                            starvation &   2150 &                 5 &             10 \\
   &   &                                  ogre &   2095 &                 5 &              9 \\
   &   &                                 rothe &   2084 &                 6 &              8 \\
   &   &  Uruk-hai called Haiaigrisai of Aruka &   2036 &                 5 &              6 \\
   \cmidrule{2-6}
   &  wiz-elf-cha-mal &                           cave spider &   1662 &                 2 &              7 \\
   &   &                                iguana &   1332 &                 2 &              5 \\
   &   &                            starvation &   1329 &                 1 &              5 \\
   &   &                            starvation &   1311 &                 1 &              5 \\
   &   &                            gnome lord &   1298 &                 5 &              9 \\
\bottomrule
\end{tabular}
}
\label{tab:results:top}
\end{table}

\begin{figure}[t!]
    \centering \includegraphics[width=\columnwidth]{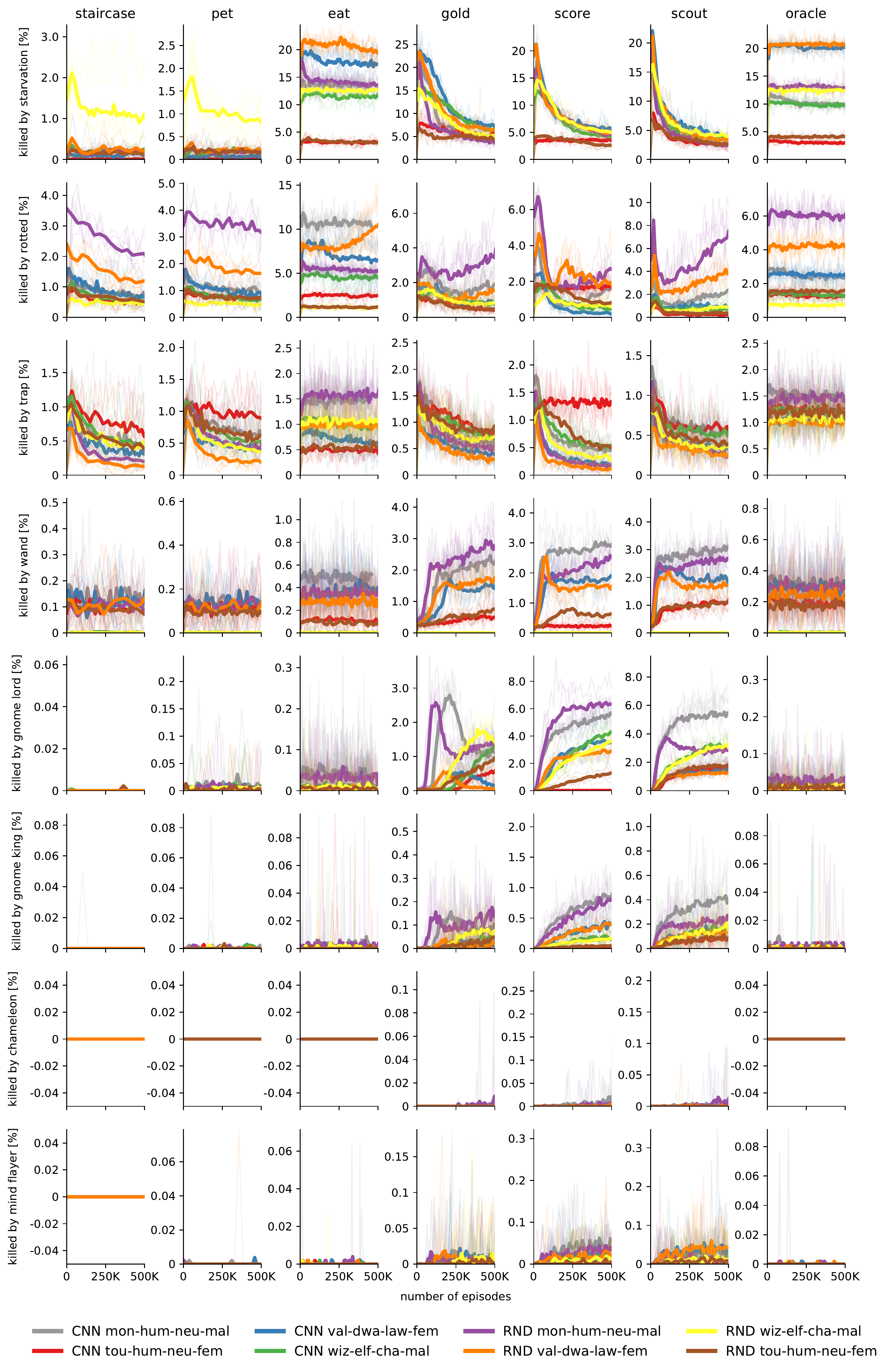}
    \caption{Analysis of different causes of death during training, averaged over the last $1000$ episodes and over five runs.}
    \label{fig:death_analysis}
\end{figure}

\begin{sidewaysfigure}[ht]
\centering
    \includegraphics[width=0.9\linewidth]{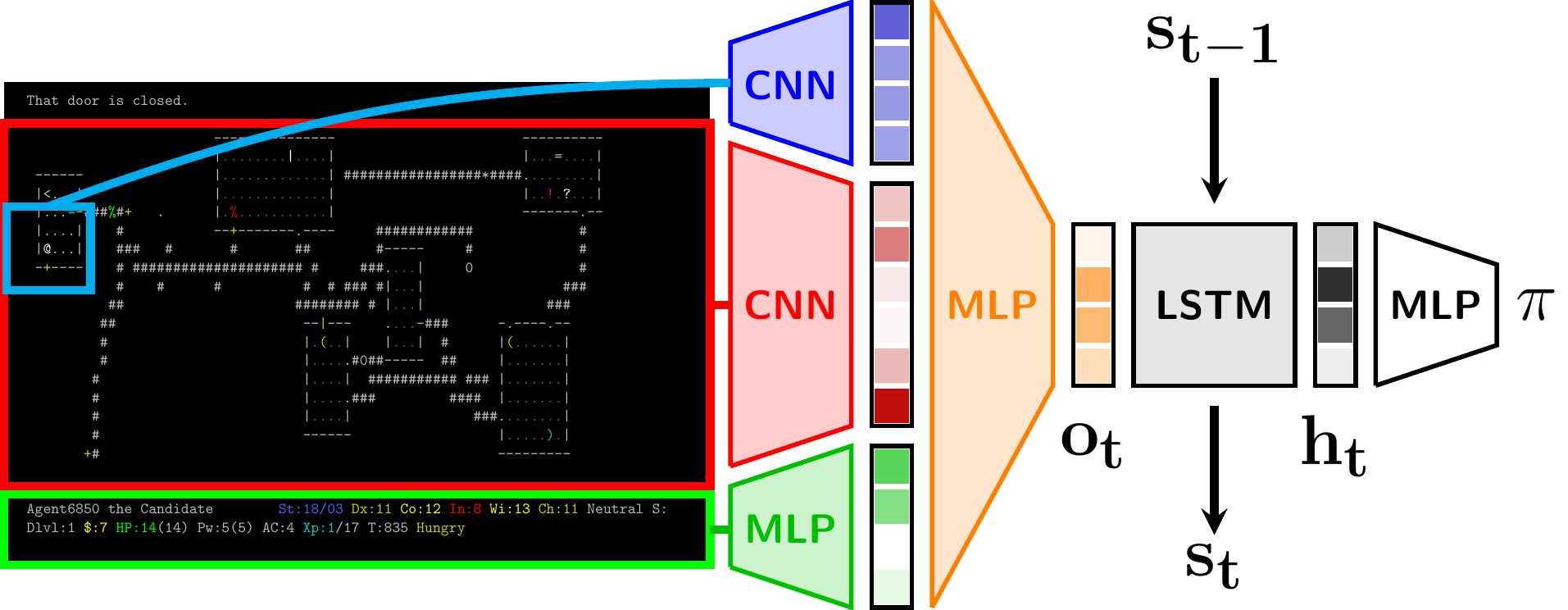}
    \caption{Overview of the core architecture of the baseline models released with \NLE{}.}
    \label{fig:modelBig}
\end{sidewaysfigure}

\end{document}